\documentclass[10pt,twocolumn,letterpaper]{article}

\usepackage[pagenumbers]{cvpr} % To force page numbers, e.g. for an arXiv version

\usepackage{xcolor}

\definecolor{cvprblue}{rgb}{0.21,0.49,0.74}
\usepackage[pagebackref,breaklinks,colorlinks,allcolors=cvprblue]{hyperref}

\usepackage{algorithm}
\usepackage{algorithmic}
\usepackage{tabularx}
\usepackage{multirow}

\begin{document}

\title{
%Dual Stereo Exposure for Adaptive-dynamic-range 3D Imaging
% Learned Dual-exposure Control for HDR 3D Imaging
% Stereoscopic HDR 3D Imaging with Learned Dual-exposure Control
%Learning Dual-exposure Control for HDR 3D Imaging
% Automatic Dual-exposure Control for Ultra-wide Dynamic Range 3D Imaging
% Automatic Dual-exposure Control for Extended Dynamic Range 3D Imaging
Dual Exposure Stereo for Extended Dynamic Range 3D Imaging
}

% \author{
% Juhyung Choi \\
% POSTECH\\
% \and
% Jinnyeong Kim \\
% POSTECH\\
% \and
% Seokjun Choi \\
% POSTECH\\
% \and
% Jinwoo Lee \\
% KAIST\\
% \and
% Samuel Brucker \\
% Torc Robotics\\
% \and
% Mario Bijelic \\
% Princeton University\\
% \and
% Felix Heide \\
% Princeton University\\
% \and
% Seung-Hwan Baek \\
% POSTECH 
% }

\author{Juhyung Choi\footnotemark[1] ~ ~ ~
Jinnyeong Kim\footnotemark[1] ~ ~ ~
Seokjun Choi\footnotemark[1] ~ ~ ~
Jinwoo Lee\footnotemark[2] ~ ~ ~ \\
Samuel Brucker\footnotemark[3] ~ ~ ~
Mario Bijelic\footnotemark[4] ~ ~ ~
Felix Heide\footnotemark[4] ~ ~ ~
Seung-Hwan Baek\footnotemark[1] \\
\footnotemark[1]~ POSTECH ~ ~ ~ \footnotemark[2]~ KAIST ~ ~ ~ \footnotemark[3]~ Torc Robotics ~ ~ ~ \footnotemark[4]~ Princeton University \\
}

\twocolumn[{
\renewcommand\twocolumn[1][]{#1}
\maketitle
\begin{center}
\vspace{-10mm}
    \centering
    \captionsetup{type=figure}
    \includegraphics[width=\linewidth]{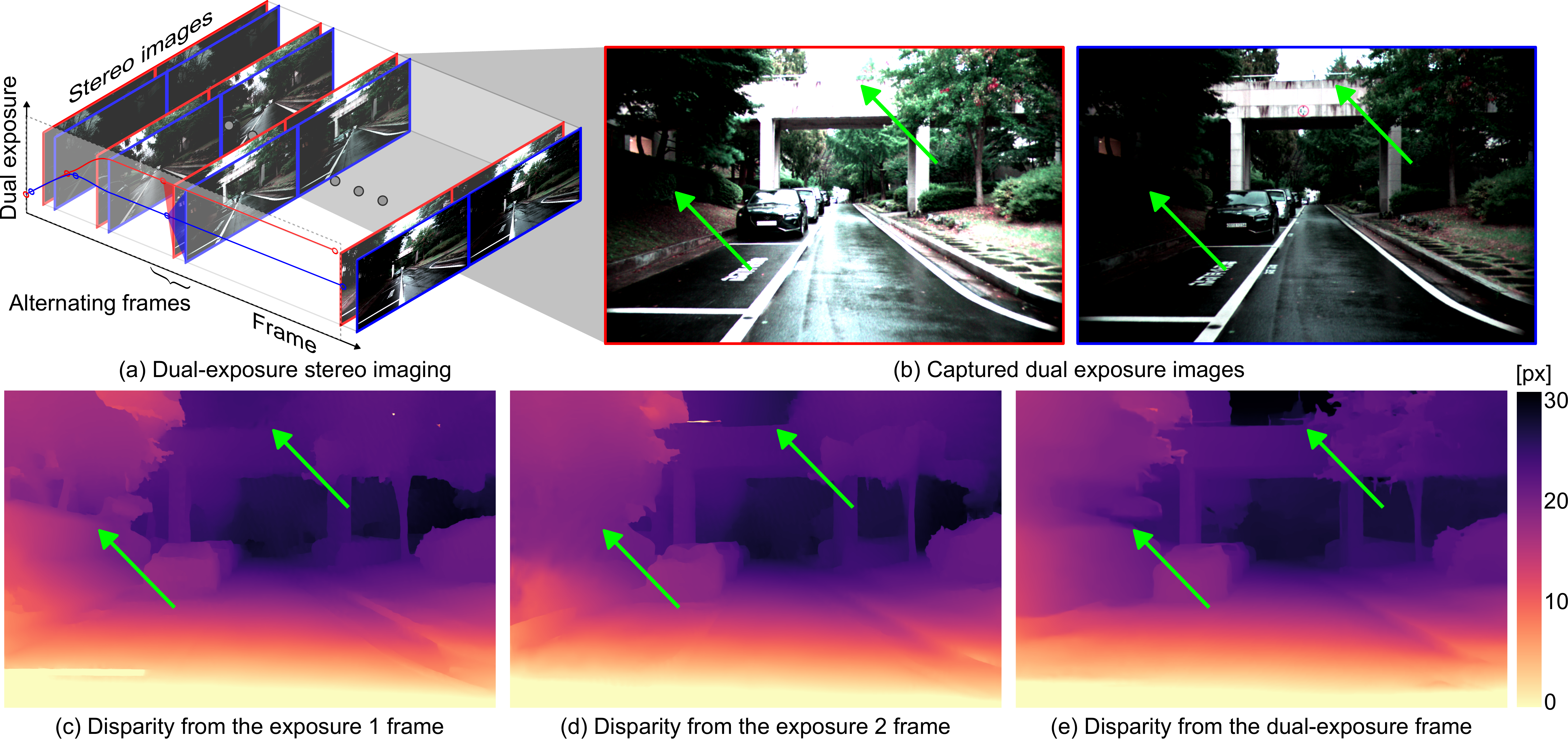}
    \vspace{-4mm}
    \captionof{figure}{We introduce dual-exposure stereo, a method for extended dynamic range (DR) 3D imaging. (a) We control the dual exposures synchronously set for the stereo camera to expand the effective DR of 3D imaging.
    From (b) the captured dual-exposure stereo images, {we estimate (e) a disparity map that preserves details in both the under- and over-exposed images of the dual-exposure pair, which cannot be faithfully reconstructed in (c)\&(d) one-exposure results.
    }
}
    \label{fig:teaser}
\end{center}
}]

\begin{abstract}
\vspace{-12mm}
Achieving robust stereo 3D imaging under diverse illumination conditions is an important however challenging task, due to the limited dynamic ranges (DRs) of cameras, which are significantly smaller than real world DR. As a result, the accuracy of existing stereo depth estimation methods is often compromised by under- or over-exposed images.
Here, we introduce dual-exposure stereo for extended dynamic range 3D imaging. We develop automatic dual-exposure control method that adjusts the dual exposures, diverging them when the scene DR exceeds the camera DR, thereby providing information about broader DR. From the captured dual-exposure stereo images, we estimate depth using motion-aware dual-exposure stereo network.
To validate our method, we develop a robot-vision system, collect stereo video datasets, and generate a synthetic dataset. Our method outperforms other exposure control methods.
\end{abstract}
\vspace{-5mm}
\section{Introduction}
\label{sec:intro}

Robust 3D imaging is critical for autonomous systems, such as robots and self-driving vehicles, which depend on depth perception to navigate and interact with their environments. Stereo imaging is a popular 3D imaging technique that estimates depth from disparity by matching corresponding pixels in images captured by two cameras. Recent advancements in neural networks have significantly improved stereo disparity estimation, making stereo imaging a practical and cost-effective solution~\cite{mayer2016large}.

However, achieving robust 3D imaging with stereo cameras remains challenging, especially in real-world scenes that exhibit lighting conditions with ultra-wide dynamic ranges (DRs). Conventional cameras have limited DR capabilities~\cite{reinhard2010high}, so in scenes with extremely wide DRs, bright regions may become overexposed while dark regions are underexposed, leading to suboptimal disparity estimation. Existing auto exposure control (AEC) methods adjust camera exposure to capture the scene DR, however they do not expand the camera’s native DR, as each stereo frame is often processed individually~\cite{cvetkovic2010automatic}. Exposure bracketing techniques capture multiple images with different exposures to expand the effective DR through multi-exposure image processing~\cite{mertens2009exposure, debevec2023recovering}, however these often rely on predefined exposures that do not adapt to the scene DR, increasing capture time and computational overhead.

In this paper, we introduce dual-exposure stereo, a method for extended dynamic range 3D imaging (Figure~\ref{fig:teaser}). In alternating frames, we capture stereo images with dual exposures, producing two pairs of stereo images where each pair is captured at different exposure settings in successive frames. By combining AEC and exposure bracketing, we dynamically adjust the dual exposures: when the scene DR exceeds the camera’s native DR, the dual exposures diverge to capture bright and dark regions across two successive frames. When the scene DR is within the camera’s DR, the exposures converge to capture the full scene DR within the camera’s native DR. Each stereo image captured under dual exposures retains the same DR as the camera, however different exposure settings enable coverage of bright and dark regions. To leverage these images, we develop a dual-exposure depth estimation method that fuses dual-exposure features in a motion-aware manner across alternating frames. Our approach effectively extends the DR for 3D imaging, regardless of the original bit depth of the cameras.

To validate our method, we design a robot-vision system equipped with stereo cameras and a LiDAR sensor. Using this setup, we collect a dataset of stereo videos and LiDAR point clouds across indoor and outdoor environments with various lighting conditions. We also generate synthetic datasets with dense ground-truth depth maps. Our experiments demonstrate that the proposed method outperforms previous exposure control methods, enabling depth estimation in scenes with a wide range of DRs. Code and datasets will be made publicly available.

In this paper, we make the following contributions:
\begin{itemize}
\item We introduce dual-exposure stereo for extended dynamic range 3D imaging, developing an automatic dual-exposure control method that combines conventional AEC and exposure bracketing. Our dual-exposure disparity estimation method then utilizes dual-exposure stereo images to increase effective camera DR for robust 3D imaging.
\item We develop a robot-vision system with stereo cameras and a LiDAR sensor mounted on a wheeled robot, collecting a real-world stereo video dataset and rendering a synthetic dataset with dense ground truth.
\item We validate our method on both synthetic and real-world datasets, demonstrating improved performance over existing exposure control methods.
\end{itemize}

\section{Related Work}
\label{sec:related}

\paragraph{HDR 3D Imaging}
Active imaging systems with engineered illumination have enabled 3D imaging under high-dynamic-range (HDR) environments.
Examples are synchronized projector-camera systems~\cite{o2015homogeneous, o20143d} and time-of-flight cameras~\cite{conde2014adaptive, shtendel2022hdr}. Without additional illumination modules, passive 3D imaging systems exploit unconventional sensors for HDR imaging, such as single-photon avalanche diodes~\cite{ingle2021passive}, quanta image sensors~\cite{fossum2016quanta, gnanasambandam2020hdr}, event cameras~\cite{zou2021learning, zhang2021event, wang2020event}, and modulo cameras~\cite{zhao2015unbounded, zhou2020unmodnet}. Using conventional cameras, capturing scenes with multiple exposures is a standard approach for HDR 3D imaging~\cite{lin2009high, chen2019new, chen2020learning}. However, these methods rely on predetermined exposure settings, which cannot adapt to changing lighting conditions, leading to loss of detail in overexposed or underexposed regions and unnecessarily long acquisition times when scene DR is low. Our method automatically adjusts dual exposure for stereo cameras, improving performance for varying-DR scenes by expanding effective camera DR for 3D imaging.

\paragraph{AEC and Exposure Bracketing}
Single-camera AEC has been extensively studied using histogram analysis~\cite{battiato2010image, schulz2007using, torres2015optimal}, model-predictive correction~\cite{vuong2008new, park2009method, su2015fast, su2016model}, entropy analysis~\cite{zhang2006automatic, ning2015optimization, lu2010novel}, and semantic analysis~\cite{yang2006face, yang2018personalized, onzon2021neural}. Extending AEC to stereo cameras, there are two common approaches. The first is to control exposure for each camera individually. However, this deteriorates stereo correspondence due to stereoscopic intensity inconsistency~\cite{kim2014auto, zhang2022automated}. The second approach is to use a synchronized AEC for stereo cameras, maintaining intensity consistency~\cite{kim2014auto, shim2018gradient}. However, existing methods in this category struggle when scene DR is larger than camera DR, leading to overexposed or underexposed regions. 
Exposure bracketing alternates multiple exposures and processes the multi-exposure images to capture a broader DR than the original camera DR~\cite{wang2018traffic, mukherjee2020backward, wang2020multi, onzon2021neural, satilmis2020per, wang2019underexposed, hold2017deep, hold2019deep}. Most existing methods use a single camera~\cite{wang2020learning, hu2022hdr} and rely on predetermined long and short exposure settings~\cite{kang2003high, kalantari2019deep, mangiat2010high, kalantari2013patch, gelfand2010multi, Chung_2023_ICCV, chen2021hdr}, which limits capture efficiency when scene DR fluctuates. 
Our method combines the principles of AEC and exposure bracketing.
We control the dual exposures of stereo cameras, thus maintaining intensity consistency between stereo images as well as expanding effective DR for 3D imaging. 

\paragraph{Stereo Depth Estimation}
Estimating depth from stereo disparity has been studied for decades~\cite{scharstein2002taxonomy}.
Recent neural-network solutions have significantly improved stereo matching by using deep feature extraction and cost volume construction~\cite{lipson2021raft, li2022practical}. These models iteratively refine disparity maps using 3D convolutions or recurrent units. 
However, they fail when scene DR is wider than camera DR, because of under- and over-exposed regions. Our disparity estimation method addresses this problem by fusing the dual-exposure images while compensating the motion between consecutive frames.

\begin{figure}[t]
  \centering
  \includegraphics[width=\linewidth]{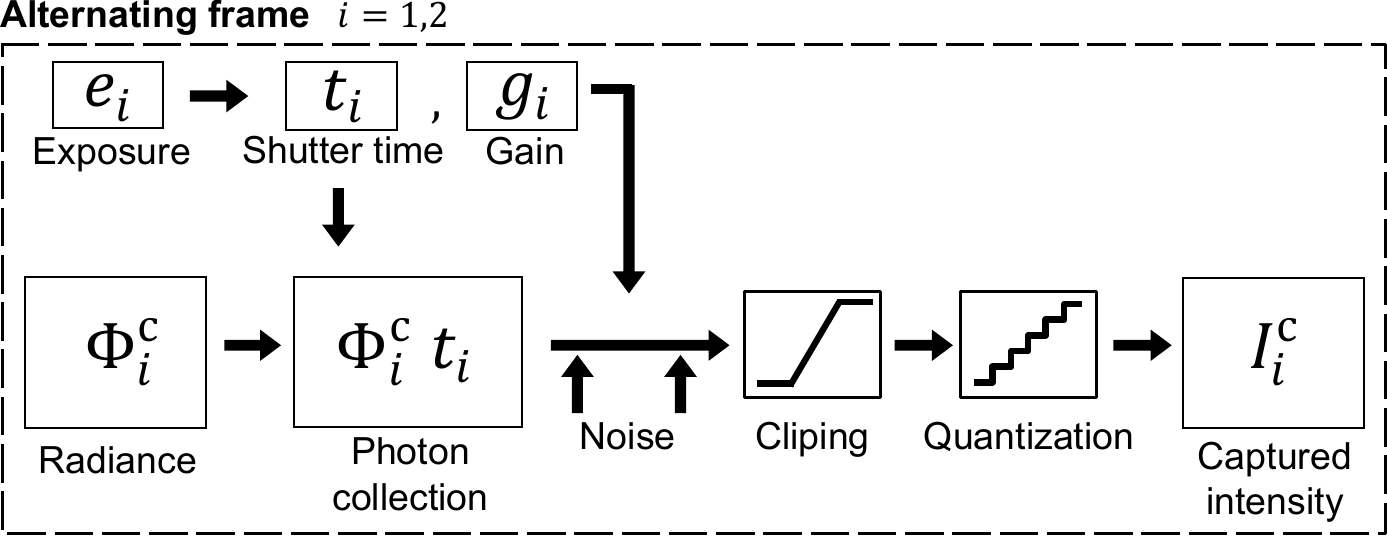}
  \caption{\textbf{Image Formation.} For alternating frame $i=\{1,2\}$, scene radiance $\Phi_i$, and exposure $e_i$, we simulate the captured intensity $I_i^c$ for the camera $c\in\{\text{left}, \text{right}\}$. We consider photon collection, pre- and post-gain noise, clipping, and quantization.} 
  \label{fig:sim}
  \vspace{-5mm}
\end{figure}

\section{Image Formation}
\label{sec:sim}
We introduce the dual-exposure image formation model for stereo cameras. 
We denote the dual exposure as $e_{i}$, where $i \in \{1,2\}$ is an alternating frame index.
We convert the exposure $e_i$ to shutter time  $t_i = e_i/g_i$ and gain $g_i = \text{max}(1,e_i/t_\text{max})$, where $t_\text{max}$ is the maximum shutter time, allocating shutter time as long as possible capped by the maximum value to reduce image noise by a high gain. 
Given the shutter time $t_i$, gain $g_i$, and the incident scene radiance $\Phi_i$, we model the intensity $I^c_i$ captured by the stereo camera $c \in \{\text{left}, \text{right}\}$ at the frame $i$ as:
\begin{equation}
\label{eq:image_formation}
I^c_{i}(p^c_{i}) = \mathrm{quant}\left(\mathrm{clip}(g_i (\Phi_i  t_i + n^\text{pre}_i ) + n^\text{post}_i)\right), 
\end{equation}
where $p^c_{i}$ is a camera pixel. 
The noise terms $n^\text{pre}_i$ and $n^\text{post}_i$ are the pre-gain and post-gain noise, sampled from zero-mean Gaussian distributions with standard deviations $\sigma_\text{pre}$ and $\sigma_\text{post}$, respectively: $n^\text{pre}_i \sim \mathcal{N}(0, \sigma_\text{pre}), \quad n^\text{post}_i \sim \mathcal{N}(0, \sigma_\text{post})$.
The function $\operatorname{clip}(\cdot)$ limits intensity values by the camera DR, and $\operatorname{quant}(\cdot)$ quantizes the intensity to integer values.
The overall procedure of the image formation is shown in Figure~\ref{fig:sim}.

\begin{algorithm}[t]
\caption{Pseudocode for ADEC.}
\label{alg:psuedo_code}
\begin{algorithmic}[1]
\REQUIRE {Dual exposure values $e_{1}, e_{2}$ and corresponding captured images $I^\text{left}_{1}, I^\text{left}_{2}$}
\ENSURE Next dual exposure $\hat{e}_{1}, \hat{e}_{2}$
\STATE \small{// Compute metrics}
\FOR{each frame $i\in\{1,2\}$}
    \STATE $h_{i}\leftarrow$ExtractHistogram$(I^\text{left}_{i})$
    \STATE $S_{i}\leftarrow$Skewness$(h_{i})$
    \STATE $L_{i},H_{i}\leftarrow$ExtremePixelRatios$(h_{i})$
\ENDFOR

\STATE \small{// Adjust dual exposure}
\IF {Scene DR $>$ camera DR: $L_{i}>\tau_{h} \text{and} H_{i}>\tau_{h}$ for any $i$}
\IF {Dual exposure gap is low: $\Delta e=|e_{1}-e_{2}| \leq \tau_{\Delta e}$}
    \STATE \small{// Diverge dual exposure}
    \STATE $\hat{e}_{1}, \hat{e}_2\leftarrow$ DivergeDualExposure$(e_{1},e_2,L_{1},L_2, H_{1},H_2)$
\ENDIF
\ELSE 
\STATE \small{// Scene DR $>$ camera DR or scene DR is uncertain}
\FOR{\normalsize{each frame $i\in\{1,2\}$}}
    \STATE \small{// Adjust dual exposure towards zeroing the skewness 
    \STATE \normalsize{$\hat{e}_{i}\leftarrow$ MakeSkewnessZero$(e_{i},S_{i})$}}
\ENDFOR
\ENDIF
\end{algorithmic}
\end{algorithm}

\begin{figure}[h]
  \centering
  \includegraphics[width=\linewidth]{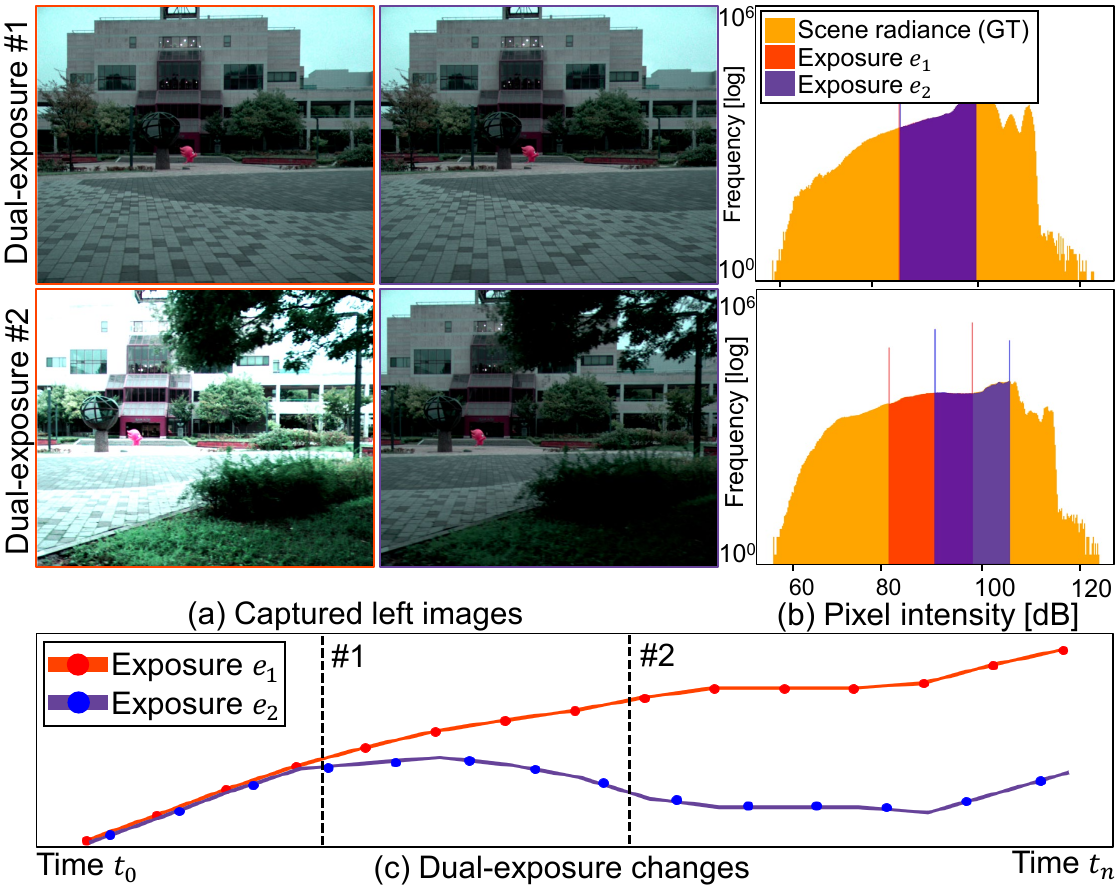}
  \caption{
  \textbf{ADEC Method Exposure Selection.} 
  {The initial dual exposures are set to be the same in this example. If the scene DR is estimated as wider than the camera DR, the dual exposures diverge to capture both dark and bright regions in alternating frames. 
  } 
  }
  \label{fig:exposure_control}
  \vspace{-5mm}
\end{figure}

\begin{figure*}[t]
  \centering
  \includegraphics[width=0.90\textwidth]{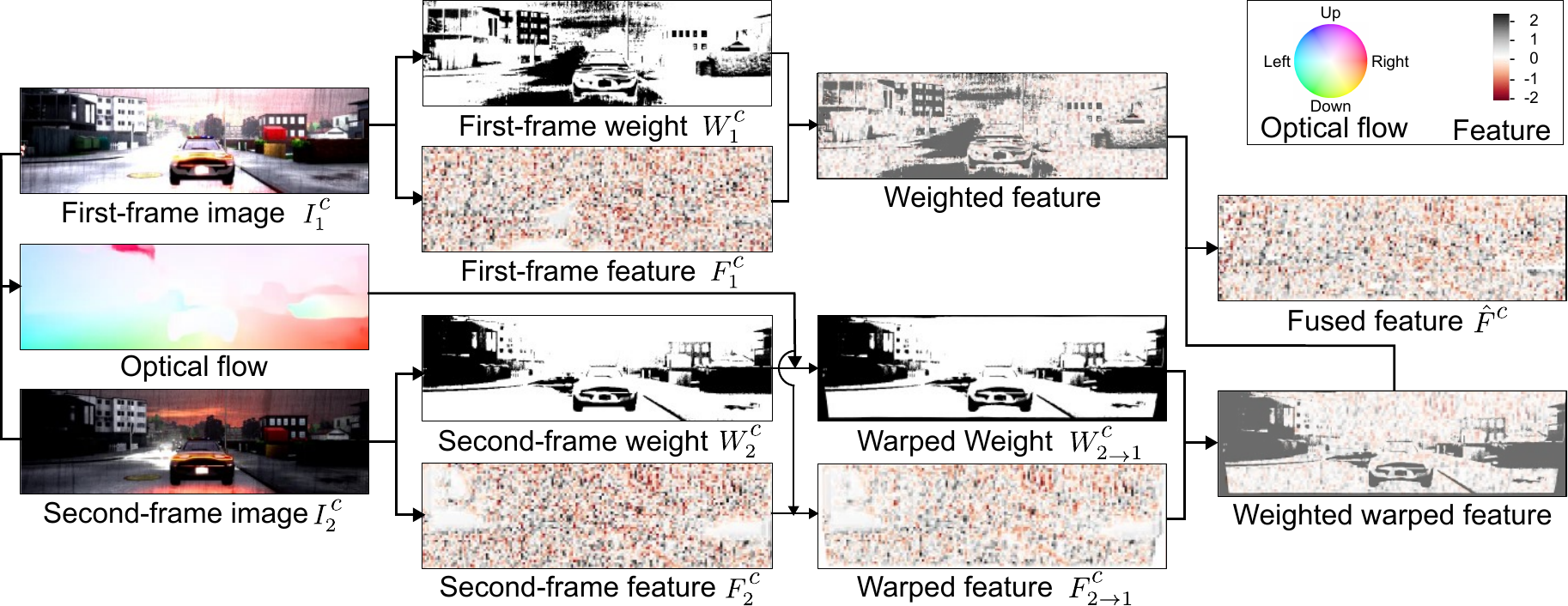}
  \caption{\textbf{Dual Exposure Feature Fusion.} For each camera $c$, we extract features $F^c_1$ and $F^c_2$, optical flow $f^c$, and weight maps $W^c_1$ and $W^c_2$ of the dual-exposure images $I^c_1$ and $I^c_2$.
  The second-frame features and weight map are warped to the first-frame view, and fused to the final feature $\hat{F}^c$ with the weighted summation, encoding dual-exposure information.
  }
  \label{fig:fusion}
  \vspace{-5mm}
\end{figure*}
\section{Auto Dual Exposure Control}
\label{sec:auto}

Our ADEC method uses the left-camera images $\{I^\text{left}_1,I^\text{left}_2\}$ captured by the dual exposure $\{e_1, e_2\}$ to estimate the next dual exposure $\{\hat{e}_1, \hat{e}_2\}$.
Pseudo code of our ADEC method is shown in Algorithm~\ref{alg:psuedo_code}.
Figure~\ref{fig:exposure_control} shows an example scenario of applying the ADEC method.

\paragraph{Metric}
Our ADEC method uses statistical metrics to control next-frame dual exposures.
Specifically, for each frame $i\in\{1,2\}$, we compute the intensity histogram $h_i$ of the image $I^\text{left}_i$, and calculate the histogram skewness $S_i$, describing whether the histogram is skewed towards low intensity (negative skewness) or high intensity (positive skewness) as
\begin{equation}
S_i = \sum_{ j=0}^{K} \left(\frac{j - K/2}{K/2}\right)^3\frac{  h_i\left(j\right)}{N},
\label{eq:histogram skewness}
\end{equation}
where $K$ is the maximum detectable intensity of the camera depending on its bit depth, $N$ is the number of pixels, and $h_i(j)$ is the frequency of intensity $j$.

We then compute the ratios of under- and over-exposed pixels, denoted as $L_i$ and $H_i$, representing the proportions of pixels near the minimum and maximum intensities
\begin{equation} L_{i} = \sum_{j=0}^{T_{\text{low}}} \frac{h_i(j)}{N}, \quad H_i = \sum_{j=T_{\text{high}}}^{K} \frac{h_i(j)}{N},
\end{equation}
where $T_{\text{low}} = \lfloor K \times 0.05 \rfloor$ and $T_{\text{high}} = \lfloor K \times 0.95 \rfloor$ are clamping thresholds.

Below, we use the skewness $S_i$, extreme-valued pixel ratios $L_i, H_i$ to determine the next-frame dual exposures.

\paragraph{Diverging Dual Exposure}
For scenes whose DR exceeds the camera native DR, we diverge the dual exposure to capture a broader DR across the alternating frames. Such cases are identified with the conditions $L_i > \tau_h$ and $H_i > \tau_h$ for at least one frame $i$, where $\tau_h = 0.05$. We diverge the dual exposure with magnitudes proportional to $L_i$ and $H_i$ as
\begin{align}
\label{eq:diverging}
    \hat{e}_{1} &= e_{1} +  \alpha L_1, \quad \hat{e}_{2} = e_{2} -  \alpha H_2, \quad \text{if} \quad e_1 > e_2  \nonumber \\
    \hat{e}_{1} &= e_{1} -  \alpha H_1, \quad \hat{e}_{2} = e_{2} +  \alpha L_2, \quad \text{otherwise,} 
\end{align}
where $\alpha = 0.5$ is a constant controlling the divergence step. 

To prevent excessive divergence that could lead to unstable stereo imaging, we limit the exposure difference $\Delta e = | e_1 - e_2 |$. If $\Delta e$ exceeds a threshold $\tau_{\Delta e}=2.5$, no further divergence is applied.

\paragraph{Towards Zeroing Skewness}
When the scene DR is uncertain or lower than the camera DR, indicated by $L_i < \tau_h$ and $H_i < \tau_h$ for both frames $i\in\{1,2\}$, we adjust the exposures towards zeroing the skewness $S_i$.
This makes the intensity histogram to be balanced, capturing the scene DR:
 \begin{equation}
    \hat{e}_{i} = {e}_{i} - \alpha \times S_{i}.
    \label{eq:adjustment_skewness}
\end{equation}

This process makes the dual exposure converge to a similar value, to fully cover the scene DR in a balanced manner, if the scene DR is lower than camera DR. For scenes with uncertain DR compared to the camera DR, this method moves the dual exposure to be with zero skewed, facilitating better identification of the scene DR.

\begin{figure*}[t]
  \centering
  \includegraphics[width=\textwidth]{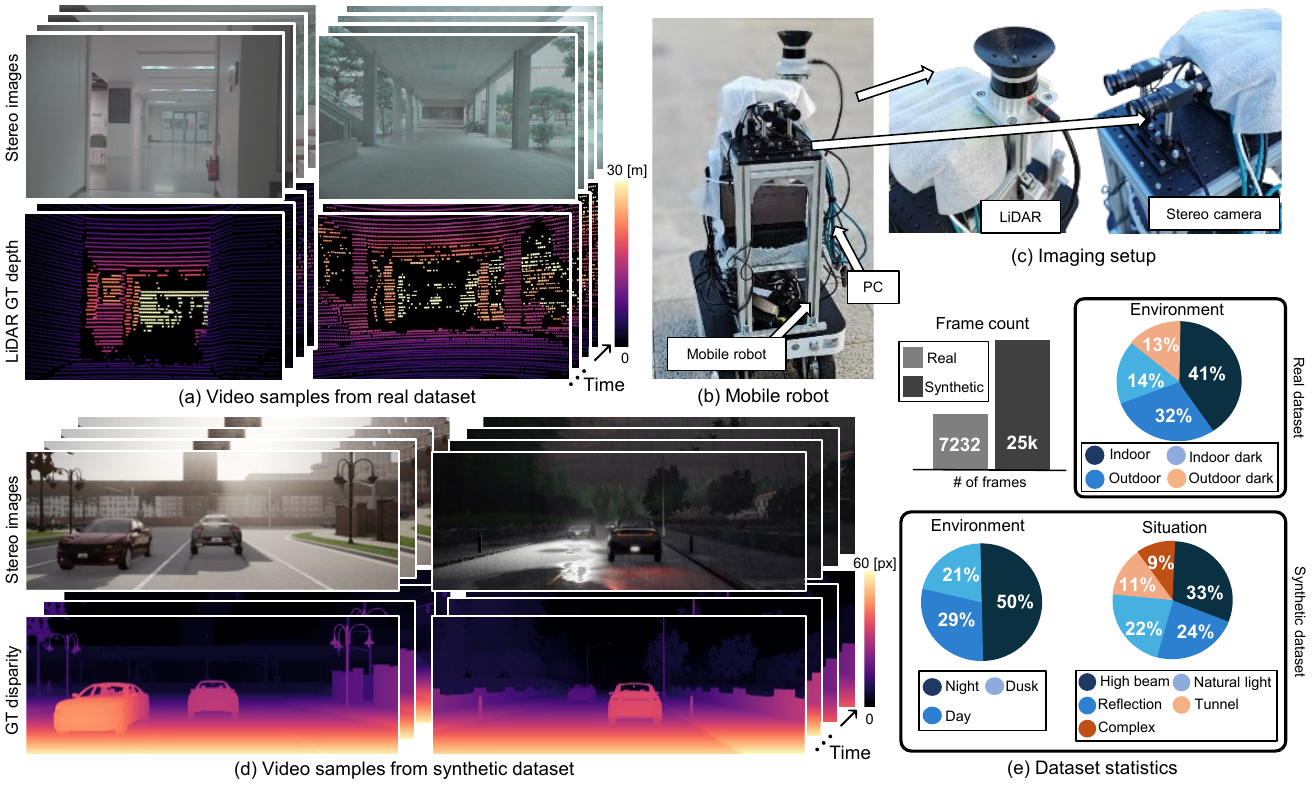}
  \caption{\textbf{Stereo Video Datasets.} (a) Samples of our real-world dataset: sequence of stereo images and corresponding LiDAR point cloud. (b)\&(c) Our imaging setup (c) mounted on a mobile robot (b), equipped with a LiDAR sensor and stereo cameras. (d) Samples of our synthetic dataset: stereo images and ground-truth disparity maps. (e) Statistics of the synthetic dataset about time, location, and scenarios.}
  \label{fig:dataset_composition}
  \vspace{-5mm}
\end{figure*}

\section{Dual-exposure Stereo Disparity Estimation} 
We introduce our disparity estimation method for dual-exposure stereo images: $I_1^\text{left}$, $I_1^\text{right}$ of the first frame, and $I_2^\text{left}$, $I_2^\text{right}$ of the second frame.

\paragraph{Motion Estimation}
Between alternating frames $i \in \{1,2\}$, the locations of corresponding pixels can move for dynamic movements of objects or cameras, modeled as the optical flow $f^c$:
\begin{align}
\label{eq:disparity_temporal}
p^c_{1} = p^c_{2} + f^c(p^c_{2}).
\end{align}
To account for such motion, we estimate the optical flow $f^\text{left}$ between $I_1^\text{left}$ and $I_2^\text{left}$, and  $f^\text{right}$ between $I_1^\text{right}$ and $I_2^\text{right}$ using a pretrained optical flow network robust to intensity difference between images~\cite{morimitsu2024rapidflow}. 
Note that the dual-exposure images $I_1^c$ and $I_2^c$ have overlapped contents as we do not allow for extremely-diverged dual exposures as described in Section~\ref{sec:auto}. 

The estimated optical flow  $f^c$ allows us to define a warping function that transforms data $\mathcal{X}_2^c$ from the second frame ($i=2$) to the first frame ($i=1$):
\begin{equation} 
X_{2 \rightarrow 1}^c = \mathrm{warp}(X_2^c, f^c),
\end{equation}
where $X_{2 \rightarrow 1}^c$ is the warped data. The warping function entails resizing the optical flow field to the corresponding resolution of the input data $X_2^c$.

\paragraph{Dual-exposure Feature Fusion}
To exploit dual-exposure images $\{I_1^c, I_2^c\}$ with potentially different exposures, we develop a dual-exposure feature fusion method.
Figure~\ref{fig:fusion} shows the overview of our dual-exposure feature fusion.
We extract features $F_i^c$ from images $I_i^c$ using a pretrained feature extractor $\text{FE}(\cdot)$~\cite{lipson2021raft}:
\begin{equation}
F_i^c = \text{FE}(I_i^c).
\end{equation}
We then warp the second-frame feature to the first frame using the warping function based on the estimated optical flow, ensuring that feature from the second frame becomes spatially aligned with the first-frame feature:
\begin{equation}
F_{2 \rightarrow 1}^c = \mathrm{warp}(F_2^c, f^c).
\end{equation}

As we now have the dual exposure features $F_{2 \rightarrow 1}^c $ and $F_1^c$ spatially aligned, we fuse the two features using the weighted sum:
\begin{equation} 
\label{eq:fusion}
\hat{F}^c = \frac{W_1^c \cdot F_1^c + W_{2 \rightarrow 1}^c \cdot F_{2 \rightarrow 1}^c}{W_1^c + W_{2 \rightarrow 1}^c + \epsilon}, 
\end{equation}
where $\hat{F}^c$ is the fused feature, $\epsilon$ is a small constant to avoid division by zero. 

We define the weight map $W_i^c$ using an intensity-based trapezoidal function~\cite{kalantari2013patch} to exploit well-exposed pixel intensity:
\begin{equation}
W_i^c(p_i^c) = 
\begin{cases}
    \frac{I_i^c(p_i^c)}{\alpha} & \text{if } I_i^c(p_i^c) < \alpha, \\
      1 & \text{if } \alpha \leq I_i^c(p_i^c) \leq \beta \\
      1 - \frac{1}{1 - \beta} \left(I_i^c(p_i^c) - \beta\right) & \text{if } I_i^c(p_i^c) > \beta,
\end{cases}
\end{equation}
where $\alpha=0.02$ and $\beta=0.98$ are the thresholds.
The weight maps $W_1^c$ and $W_{2 \rightarrow 1}^c$ are computed for the first frame and the second frame followed by being warped to the first-frame using the estimated optical flow.

We apply the dual-exposure feature fusion of Equation~\eqref{eq:fusion}, obtaining the fused feature maps $\hat{F}^\text{left}$ and $\hat{F}^\text{right}$.

\begin{figure*}[t]
  \centering
  \includegraphics[width=\textwidth]{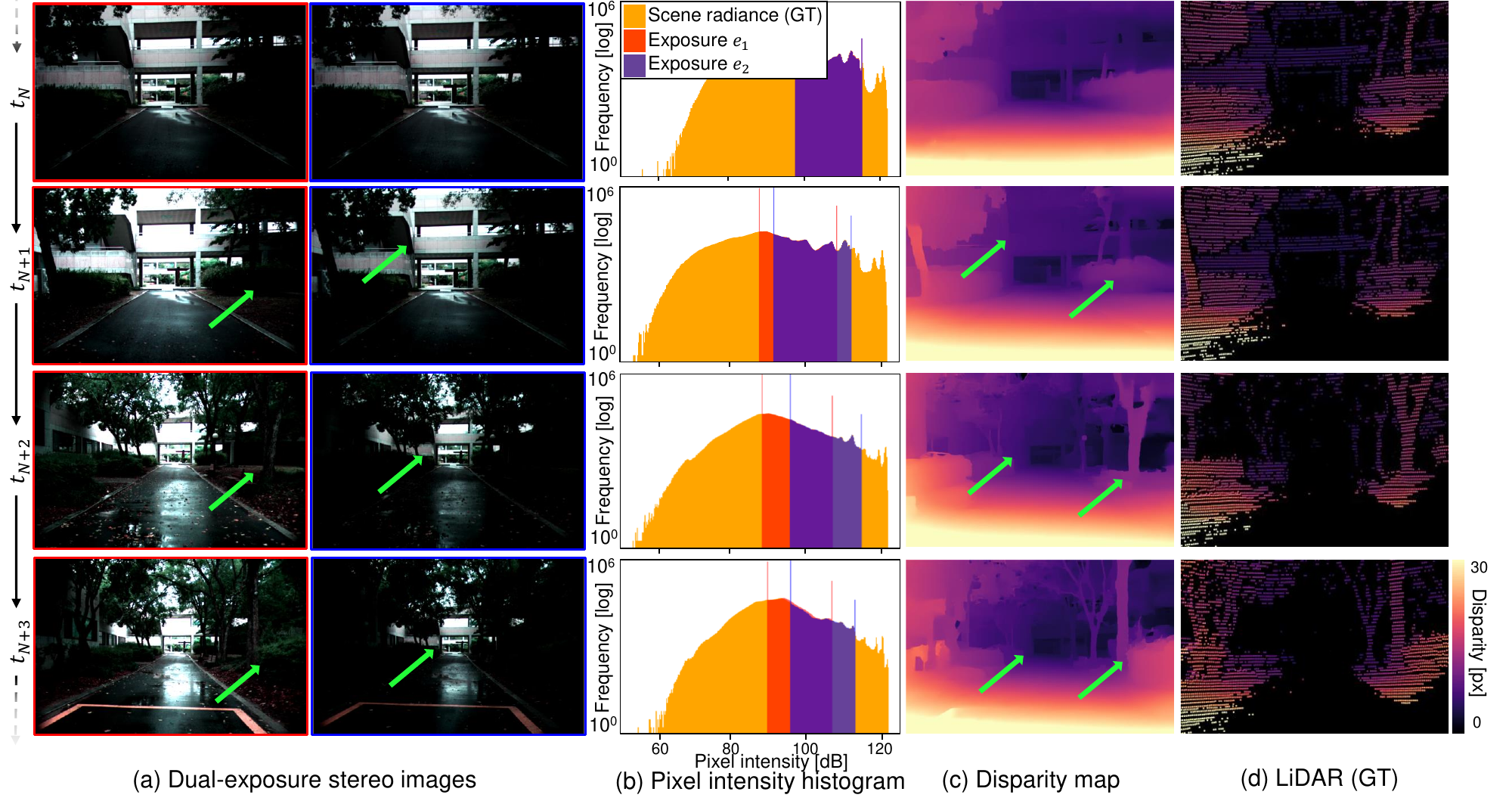}
  \caption{\textbf{Extended-DR 3D imaging.} Our ADEC controls the dual exposures for extended-DR depth estimation. The dual-exposure depth estimation module obtains depth both for bright and dark regions which cannot be captured by only one image. }
  \vspace{-5mm}
  \label{fig:result}
\end{figure*}

\paragraph{Stereo Disparity Estimation}
Using the fused feature maps, we construct correlation volumes $C$ as 
\begin{equation}
        C(x,y,d) = \hat{F}^{\text{left}}(x,y)\cdot \hat{F}^{\text{right}}(x+d,y),
\end{equation}
where $x,y$ are the pixel coordinates and $d$ is the disparity.
Our dual-exposure feature fusion encodes both dark and bright features of dual-exposure stereo images in the correlation volume, allowing for effectively extended DR for 3D imaging.
We also apply a multi-scale feature fusion approach for robustness. 
We then estimate a disparity map from the correlation volume using a disparity estimation network~\cite{lipson2021raft}. 
We finetune the network on our synthetic dataset (Section~\ref{sec:synthetic dataset}).

\section{Stereo Video Datasets}
\label{sec:dataset}
To validate our method, we introduce two stereo video datasets: one for real-world scenes and the other for synthetic scenes. 
Figure~\ref{fig:dataset_composition} visualizes samples from our datasets.

\subsection{Real-world Dataset}

\paragraph{Prototype System}
We developed an imaging system consisting of stereo RGB cameras (LUCID Triton 5.4MP) and a LiDAR sensor (Ouster OS-1). The cameras capture linear stereo images with 24-bit depth, which are then compressed to simulate 8-bit images using Equation~\eqref{eq:image_formation} to evaluate whether our method can expand effective DR for 3D imaging. Note that our method can be applied to cameras with any bit depth. {We performed intrinsic and extrinsic calibration of the stereo camera system using a chessboard-based method. Subsequently, we estimated the extrinsic transformation between the LiDAR and the left camera through ICP alignment process.} LiDAR point clouds are projected onto the {left-camera view}, providing pseudo ground-truth sparse depth maps. The stereo images and LiDAR depth maps are time-synchronized. We mount the imaging system on a wheeled robot (AgileX Ranger-Mini 2.0), as shown in Figure~\ref{fig:dataset_composition}(c), enabling both indoor and outdoor captures. 
{We configured a stereo camera system with a 110 mm baseline to enable effective depth estimation in both indoor and outdoor environments. The depth measurement range of our system extends up to 60 meters.} For system details, we refer to the Supplementary Document.

\paragraph{Captured Dataset}
Our real-world dataset encompasses a broad range of environments and lighting conditions, thus appropriate for evaluating our method. The dataset includes 33 scenes and 7432 frames, with a resolution of 1440$\times$928 pixels. The dataset is balanced with 41\% of frames captured in indoor scenes, 32\% in outdoor scenes, 14\% in indoor low-light scenes, and 13\% in outdoor low-light scenes. Further details are provided in the supplementary material.

\subsection{Synthetic Dataset}
\label{sec:synthetic dataset}
We use the CARLA simulator~\cite{dosovitskiy2017carla} to generate a synthetic dataset for diverse automotive scenarios with varying lighting conditions. We render synchronized stereo images with 32 bit depth, which is used to simulate 8-bit images using Equation~\eqref{eq:image_formation}.
We also render ground-truth dense disparity maps.
Our synthetic dataset comprises 1,000 training videos and 200 testing videos. Training videos consist of 20 frames each, and testing videos contain 100 frames each. We simulate various lighting conditions (day, dusk, and night), as shown in Figure~\ref{fig:dataset_composition}, which vary within each video, introducing abrupt changes in dynamic range due to environmental effects such as high beams at night, intense reflections, and sunlight emerging from tunnels
Further details are provided in the Supplementary Material.

\section{Experiments}
\label{sec:assessment}

\begin{figure}[t]
    \centering
    \includegraphics[width=\linewidth]{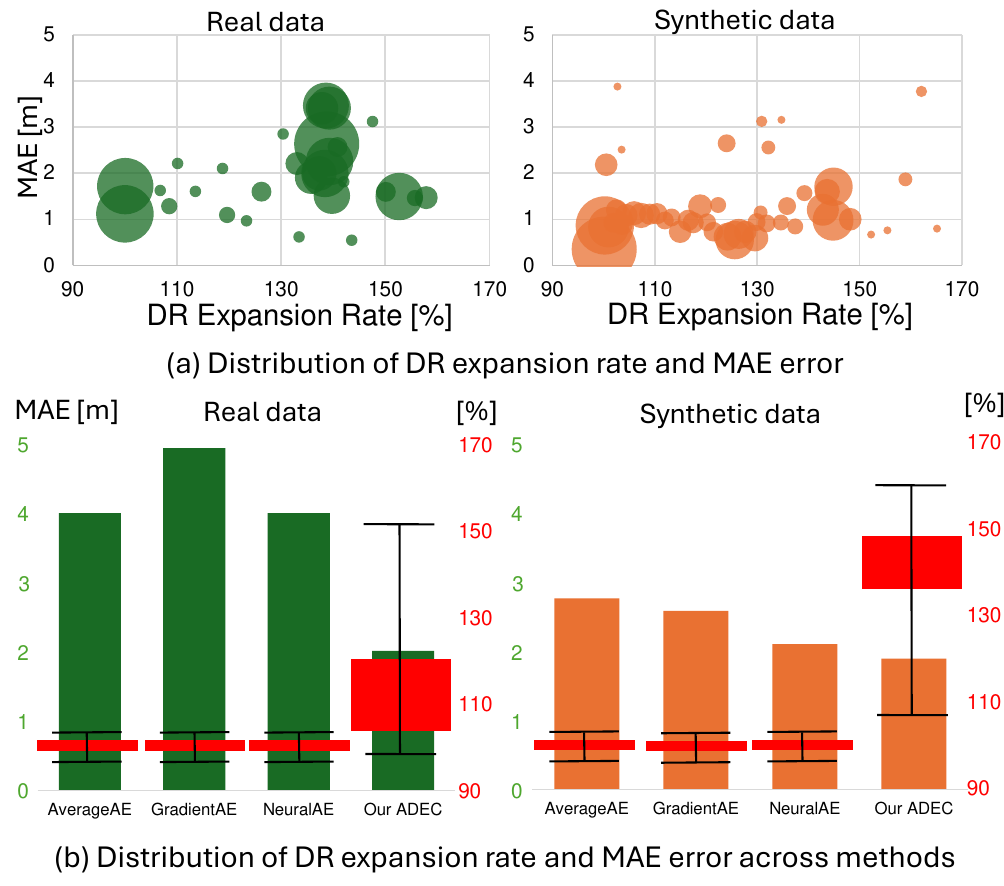}
    \caption{\textbf{Disparity accuracy vs. DR expansion rate.} (a) We demonstrate effective DR expansion up to 160\% without significant performance drop. (b) Other AEC methods~\cite{cameraProduct,shim2018gradient,onzon2021neural} cannot expand DR resulting in large error.} 
\vspace{-5mm}
    \label{fig:tradeoff}
\end{figure}

\begin{figure}[t]
  \centering
  \includegraphics[width=\linewidth]{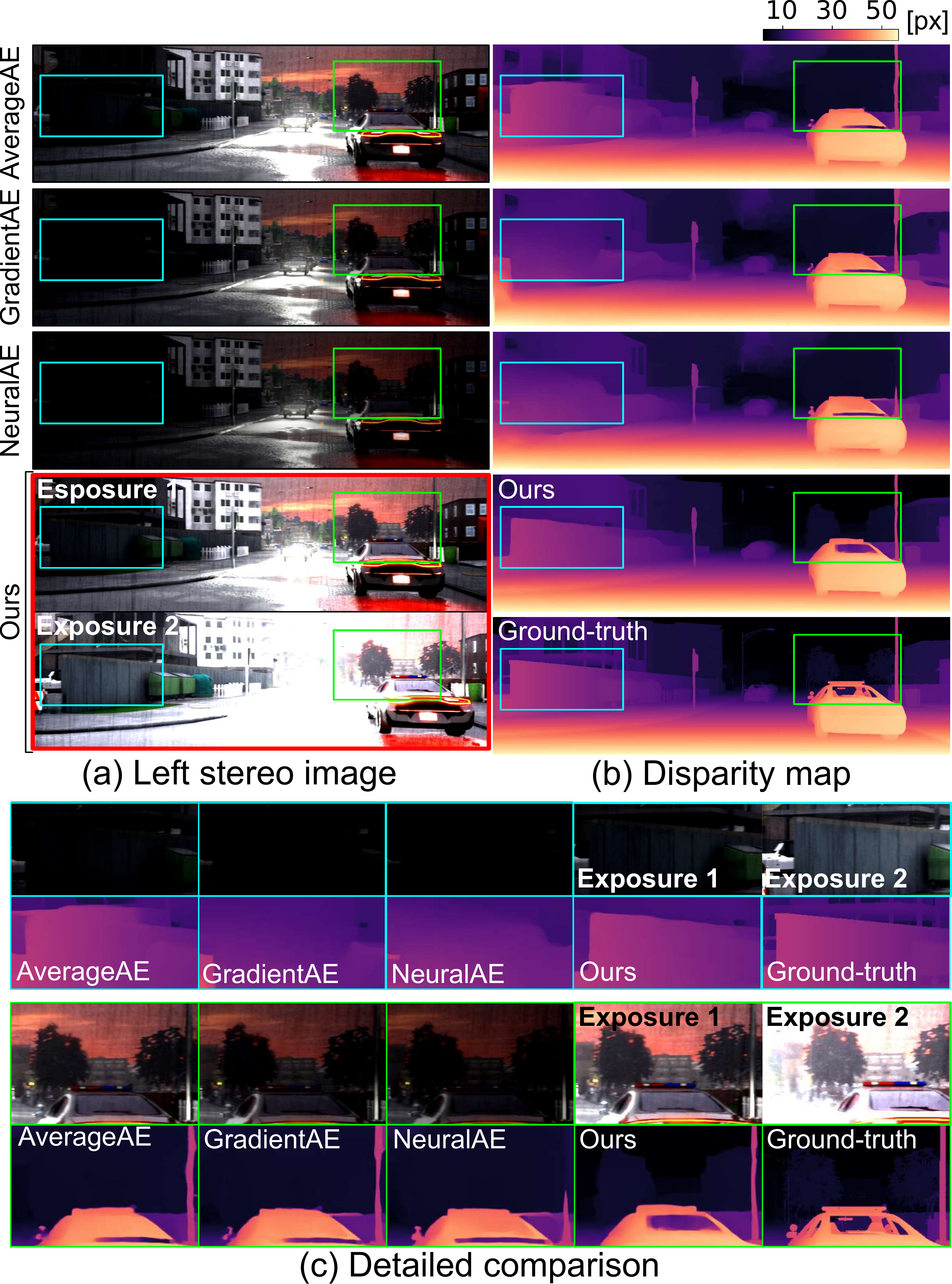}
  \caption{\textbf{Disparity accuracy using our ADEC compared with other AEC methods.}
  Our method outperforms the other AEC methods: AverageAE~\cite{cameraProduct}, GradientAE~\cite{shim2018gradient}, and NeuralAE~\cite{onzon2021neural}. 
  \label{fig:comparison_syn}
  \vspace{-3mm}
  }
\end{figure}

\begin{figure*}[t]
  \centering
  \includegraphics[width=\textwidth]{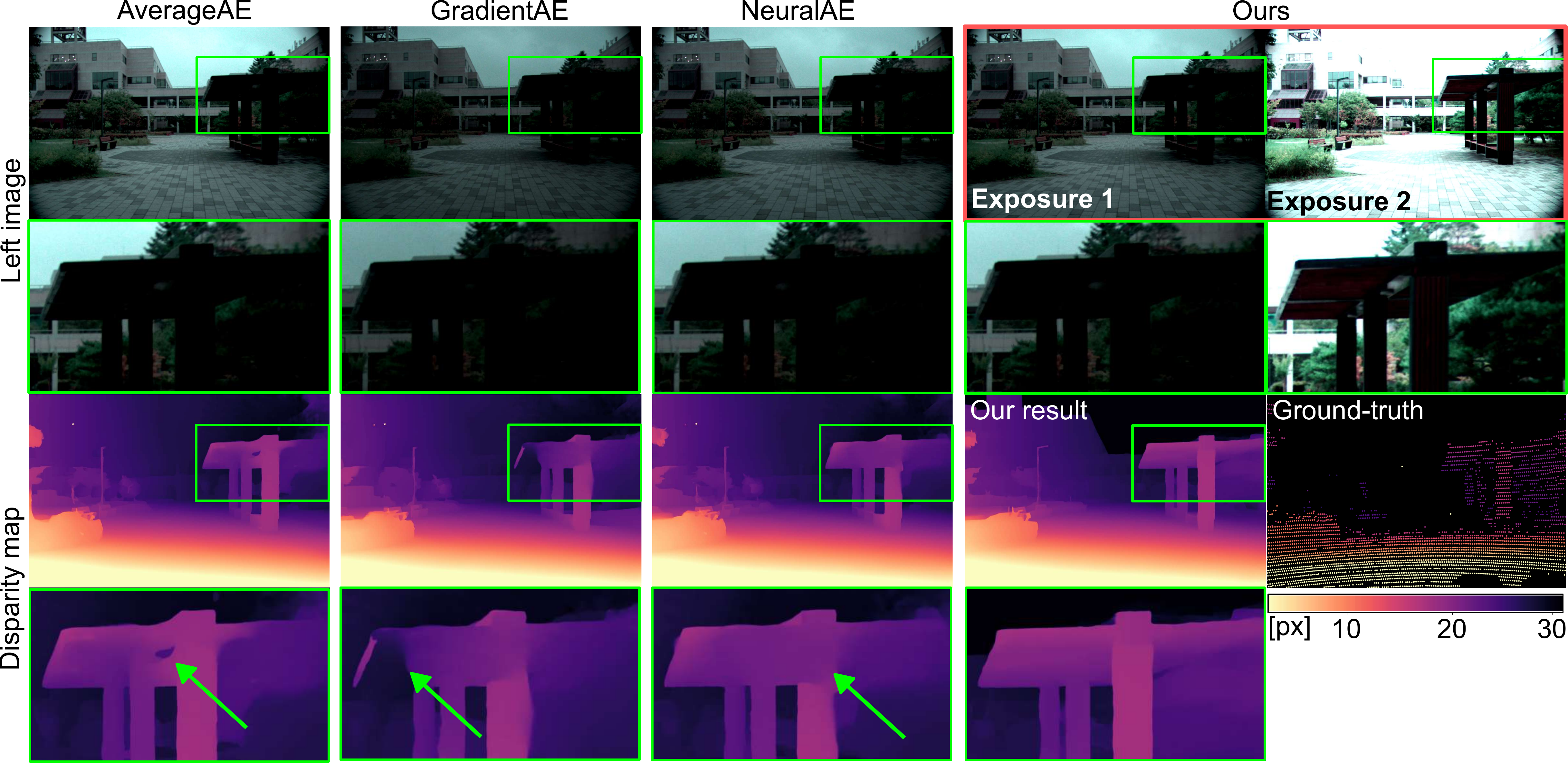}
  \caption{{\textbf{Disparity-estimation results using our ADEC compared with other AEC methods.}
  Our ADEC method outperforms the other AEC methods for subsequent extended-DR depth estimation: AverageAE~\cite{cameraProduct}, GradientAE~\cite{shim2018gradient}, and NeuralAE~\cite{onzon2021neural}. 
  }}
  \vspace{-3mm}
  \label{fig:comparison_real}
\end{figure*}

\begin{figure*}[t]
  \centering
  \includegraphics[width=\textwidth]{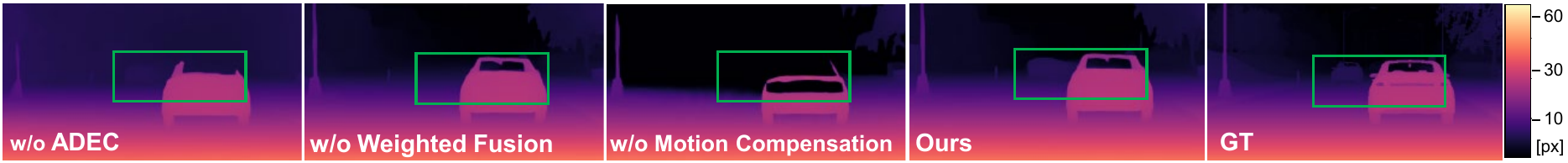}
  \caption{\textbf{Ablation study.} We evaluate the impact of three modules: ADEC, weighted feature fusion, and motion compensation. Our full method enable the most accurate reconstruction.
  }
  \vspace{-4mm}
  \label{fig:ablation}
\end{figure*}

\paragraph{Extended-DR 3D Imaging} 
We leverage complementary information of two-frame stereo images captured with dual exposure, which is estimated by our ADEC module. By fusing dual-exposure features, we obtain accurate disparity map that preserves details across a wider DR than the native camera DR.
Figure~\ref{fig:result} reports the estimated disparity maps for input dual-exposure stereo images. Our method enables expanding effective DR for disparity estimation with both details of dark and bright regions, which cannot be solely captured with the camera DR.

We assess the trade-off between the depth accuracy and the DR expansion rate.
Compared to the maximum DRs of 42dB corresponding to 8-bit depth, we calculate the effectively-enhanced DR covered by the dual-exposure frames in our method. Figure~\ref{fig:tradeoff} shows that our method retains high depth accuracy for the DR expansion rate of 160\%, which demonstrates the effectiveness of our method.
In contrast, depth estimation using other state-of-the-art AEC methods~\cite{cameraProduct,shim2018gradient,onzon2021neural} cannot expand the DR, resulting in large error in depth reconstruction. 

\paragraph{Comparison}
We compare our method with three state-of-the-art AEC methods that control and process single exposure~\cite{cameraProduct,shim2018gradient,onzon2021neural}.
We use the RAFT-Stereo model~\cite{lipson2021raft} for their depth estimation. 
Note that ours is the only method that combines the principles of AEC and exposure bracketing, which is exploited by our dual-exposure disparity estimation module.
Figures~\ref{fig:comparison_syn} and \ref{fig:comparison_real} show the qualitative results on a synthetic scene and a real-world scene, demonstrating that our method only recovers details in both dark and bright regions by effectively expanding DR for 3D imaging.
Table~\ref{tab:synthetic_comparison} confirms that our method also quantitatively outperforms the other AEC methods for accurate 3D imaging on both synthetic and real-world datasets. We also compare the speeds of AEC methods and our ADEC method considering its real-time use cases, which is an important factor for any exposure control methods. Our ADEC can run at more than 120\,FPS, supporting real-time applications, while the neural AEC~\cite{onzon2021neural} fails to support real-time imaging.

\begin{table}[t]
    \centering
    \resizebox{\linewidth}{!}{
    \begin{tabular}{c|cccc}
        \toprule[1pt]
        
             & \textbf{AverageAE} & \textbf{GradientAE} & \textbf{NeuralAE} & \textbf{ADEC} \\ 
                             & ~\cite{cameraProduct} & ~\cite{shim2018gradient} & ~\cite{onzon2021neural} &  \textbf{(ours)} \\ \hline
            \textbf{Synthetic Data} &\multirow{2}{*}{2.823} &\multirow{2}{*}{2.948} & \multirow{2}{*}{\underline{2.778}} & \multirow{2}{*}{\textbf{1.355}} \\
            \textbf{Disp. MAE [px] $\downarrow$} & & & & \\ \hline
            \textbf{Real Data} &\multirow{2}{*}{2.7679} &\multirow{2}{*}{2.5847} &\multirow{2}{*}{\underline{{1.9232}}} &\multirow{2}{*}{\textbf{1.9142}} \\
            \textbf{Depth MAE [m] $\downarrow$} & & & & \\ \hline
            \textbf{FPS$\uparrow$} & \textbf{616.27} & 42.10 & 0.25 & \underline{124.58} \\
          
        \bottomrule[1pt]  
        \end{tabular}}
        \caption{\textbf{3D imaging accuracy and FPS comparison of our ADEC against other methods.} Our method outperforms the other AEC methods in terms of disparity and depth MAE on synthetic and real datasets, respectively. We also compare FPS. Best numbers are in \textbf{bold} and the second best are in \underline{underline}.
        }
        \vspace{-5mm}
    \label{tab:synthetic_comparison}
\end{table}

\paragraph{Ablation Study}
We evaluate the importance of the three core components: ADEC module, weighted feature fusion, and motion compensation. 
Table~\ref{tab:ablation} and Figure~\ref{fig:ablation} show results.
First, instead of using our ADEC method, we fix the dual exposure to low and high values respectively using the average scene statistics.
This results in the failure of adaptation to varying scene DR, leading to high disparity error.
Second, we exclude the weighted fusion in our dual-exposure disparity estimation: we set the weight maps to be one for all pixels: $W_i^c=1$. The resulting fused features are affected by unstable features from under- or over-exposed features, leading to disparity error.
Third, we omit the motion compensation: the optical flow is set to be zero in our depth estimation process. This results in significant misalignment errors in the fused feature, making the disparity accuracy low.
Our complete method enables highest accuracy. 

\newcommand{\omark}{\text{O}} % O mark
\newcommand{\xmark}{\text{\sffamily X}} % X mark

\begin{table}[t]
    \centering
    \resizebox{0.82\columnwidth}{!}{
        \begin{tabular}{ccc|c}
        \toprule[1pt]
            \multirow{2}{*}{\textbf{ADEC}} & \textbf{Weighted } & \textbf{Motion}       & \textbf{Disparity} \\
                                           & \textbf{fusion}    & \textbf{compensation} & \textbf{MAE [px]$\downarrow$} \\
            \midrule
            $\times$ & $\checkmark$ & $\checkmark$ & 6.2775 \\ % Fixed dual exposure + Depth Estimation
            $\checkmark$& $\times$ & $\checkmark$ & 3.3968 \\ % Adaptive Dual Exposure (No Weighted Fusion)
            $\checkmark$ & $\checkmark$ & $\times$  & 8.3657 \\ % Adaptive Dual Exposure (No Flow Alignment)
            $\checkmark$ & $\checkmark$ & $\checkmark$ & \textbf{2.9010} \\ % Example row with Motion Compensation only
        \bottomrule[1pt]
        \end{tabular}
    }
    \caption{\textbf{Ablation study.} }
    \label{tab:ablation}
    \vspace{-5mm}
\end{table}

\section{Conclusion}
\label{sec:conclusion}
In this work, we introduce a dual exposure stereo for extended DR 3D imaging. We devise a ADEC module and dual-exposure depth estimation method, expanding the effective DR for robust 3D imaging. To validate this method, we report a stereo video dataset consisting of stereo videos and LiDAR pointclouds, collected by our robot vision system.
We evaluate the effectiveness of our method,  outperforming conventional AEC methods across all experimental settings we tested. As a method that can expand the DR of any stereo camera for 3D imaging, we hope the proposed approach can be a step to depth imaging in even more extreme lighting and enviromental conditions, including photon-starved captures in fog, rain, or snow.

\clearpage
%%%%%%%%% REFERENCES
{\small
\bibliographystyle{ieeenat_fullname}
\bibliography{references}
}

\end{document}

\begin{CJK}{UTF8}{mj}
% \CJKfamily{mj}

%%%%%%%%% TITLE
\title{Dual Exposure Stereo for Extended Dynamic Range 3D Imaging}

\author{
Juhyung Choi \\
POSTECH\\
\and
Jinnyeong Kim \\
POSTECH\\
\and
Jinwoo Lee \\
KAIST\\
\and
Samuel Brucker \\
Torc Robotics\\
\and
Mario Bijelic \\
Princeton University\\
\and
Felix Heide \\
Princeton University\\
\and
Seung-Hwan Baek \\
POSTECH 
}

\maketitle

In this supplemental document, we provide additional results and details in support of our findings in the main manuscript. 

\tableofcontents

\newpage

\section{Details on Image Formation}

\subsection{Image Preprocessing}
We develop a comprehensive image pre-processing pipeline. This section provides a detailed description of the pre-processing steps, including data handling, Bayer to RGB conversion, bilateral filtering, and stereo rectification.

\paragraph{Conversion from Bayer to RGB}
The raw Bayer images are first converted into 32-bit Bayer patterns, packing three 8-bit channels into a 32-bit representation. This representation is crucial for preserving the full dynamic range of the raw image data. Since the camera stores RAW image data in a custom 24-bit format, standard OpenCV functions cannot be directly applied for Debayering. To address this, we implemented a bilinear interpolation-based Debayering method. This approach reconstructs the red, green, and blue channels by interpolating the Bayer pattern, ensuring minimal color distortion. After interpolation, OpenCV’s \texttt{cvtColor} function is used to convert the interpolated Bayer image into a standard RGB format.

\paragraph{Bilateral Filtering}
To reduce grid-like artifacts introduced during Bayer to RGB conversion, bilateral filtering is applied using the OpenCV's \texttt{bilateralFilter} function. We used a spatial parameter \texttt{sigmaSpace = 20} and color parameter \texttt{sigmaColor = 20} to maintain a balance between smoothing and edge retention.

\paragraph{Stereo Image Rectification}
Accurate stereo rectification is essential for consistent disparity calculation. Using calibration data, we rectified the left and right images to align their epipolar lines. The calibration data includes intrinsic matrices, distortion coefficients, rotation, and translation parameters. Stereo rectification was performed using OpenCV’s \texttt{stereoRectify} and \texttt{initUndistortRectifyMap} functions. During rectification, the \texttt{alpha} parameter was set to 0, ensuring no blank regions were left in the rectified images by cropping areas outside the valid region. This approach produces rectified images suitable for disparity estimation with minimized distortions and artifacts.

\begin{figure}[H]
  \centering
  \includegraphics[width=\linewidth]{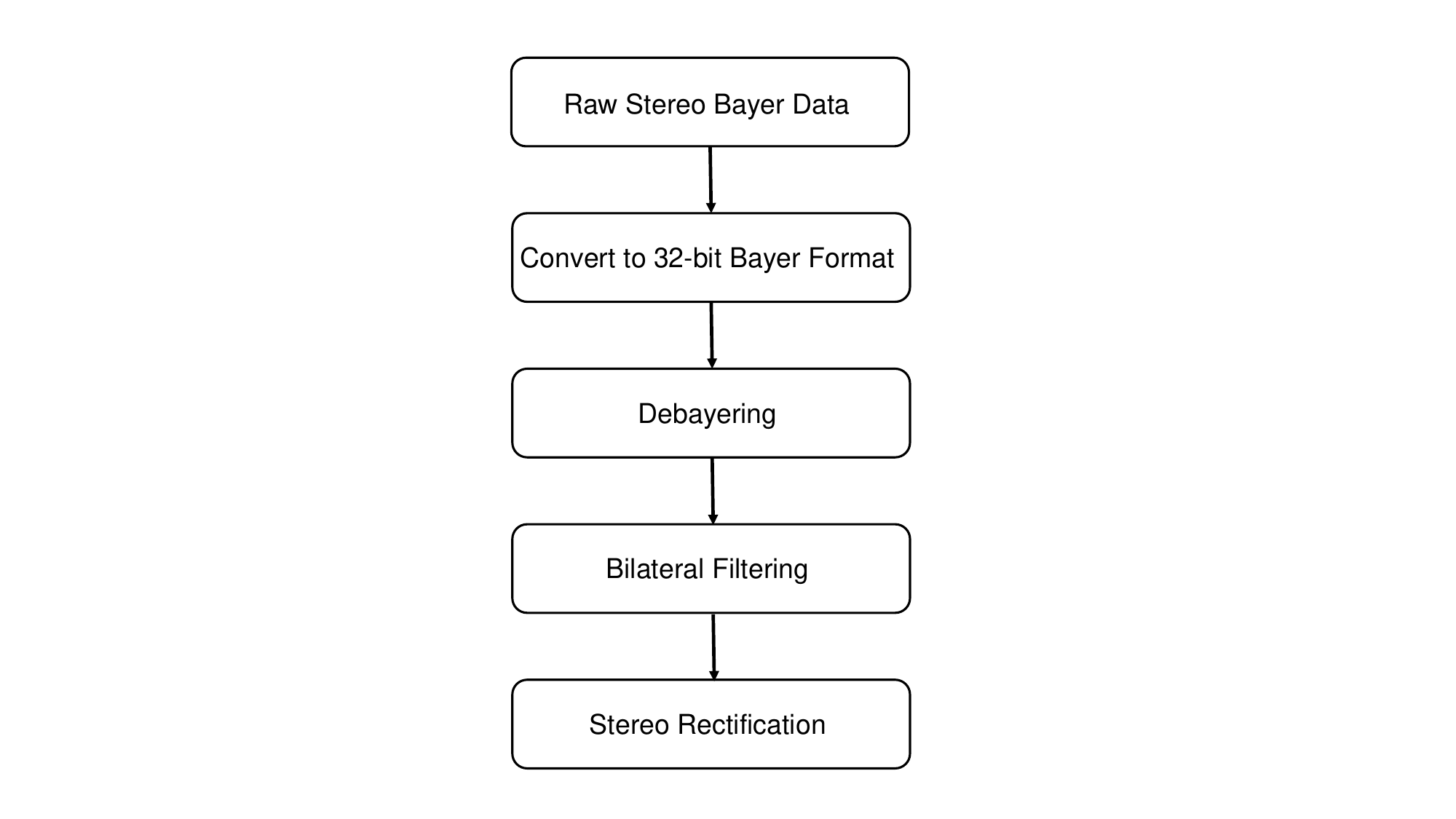}
  \caption{\textbf{Image Pre-Processing Pipeline.} The pipeline includes (1) loading raw Bayer data, (2) converting 24-bit raw Bayer patterns to 32-bit Bayer format, (3) performing debayering with custom bilinear interpolation and OpenCV color conversion, (4) applying bilateral filtering to reduce grid-like artifact, (5) rectifying stereo images using calibration parameters. This pipeline ensures high-quality and geometrically consistent inputs for disparity estimation.}
  \label{fig:preprocessing_pipeline}
\end{figure}

\paragraph{Pipeline Overview}
The pre-processing pipeline combines raw data loading, Bayer to RGB conversion, bilateral filtering, stereo rectification, and tensor conversion. These steps collectively enhance image quality and geometric consistency, enabling accurate and robust disparity estimation in subsequent stages of the pipeline. A diagram summarizing the pipeline is presented in Figure~\ref{fig:preprocessing_pipeline}.

\subsection{Image Formation}
\label{sec:Image_formation}
To simulate dual-exposure stereo image captures, we model the image formation process using exposure settings and the incident scene radiance. This process is critical for accurately simulating the captured intensity values generated under various exposures. The procedure is formalized in Equation~\eqref{eq:image_formation} and implemented in our pipeline.

\paragraph{Exposure Modeling}
We denote the exposure for each frame as \(e_i\), where \(i \in \{1,2\}\) alternates for consecutive frames. The exposure is converted to shutter time \(t_i\) and gain \(g_i\) as follows:
\begin{equation}
t_i = \frac{e_i}{g_i}, \quad g_i = \max(1, \frac{e_i}{t_\text{max}}),
\end{equation}
where \(t_\text{max}\) is the maximum allowable shutter time. This formulation ensures the longest possible shutter time is used to minimize noise, while higher gains compensate for cases where \(e_i > t_\text{max}\).

\paragraph{Noise Modeling and Clipping}
Given the incident scene radiance \(\Phi_i\), the intensity captured by camera \(c \in \{\text{left}, \text{right}\}\) at pixel \(p^c_i\) is modeled as:
\begin{equation}
\label{eq:image_formation}
I^c_i(p^c_i) = \mathrm{quant}\left(\mathrm{clip}\left(g_i (\Phi_i t_i + n^\text{pre}_i) + n^\text{post}_i\right)\right),
\end{equation}
where \(n^\text{pre}_i\) and \(n^\text{post}_i\) are pre-gain and post-gain noise terms, respectively, sampled from zero-mean Gaussian distributions:
\[
n^\text{pre}_i \sim \mathcal{N}(0, \sigma_\text{pre}), \quad 
n^\text{post}_i \sim \mathcal{N}(0, \sigma_\text{post}).
\]
The \(\mathrm{clip}(\cdot)\) function limits the intensity values within the dynamic range of the camera, defined by the bounds \([\Phi_\text{lower}, \Phi_\text{upper}]\), and \(\mathrm{quant}(\cdot)\) quantizes the intensity values to discrete levels.

\paragraph{Dynamic Range Clipping}
The simulation pipeline begins by applying the exposure settings to the reference scene radiance \(\Phi_i\), followed by noise modeling and dynamic range clipping. This process ensures that the simulated captured intensity values are consistent with the physical limitations of a camera's dynamic range.

\begin{itemize}
    \item  \textbf{Dynamic Range Initialization.} Given the scene radiance \(\Phi_i\), the dynamic range bounds are computed based on its distribution. The midpoint of the radiance, \(\Phi_\text{middle}\), is defined as:
\[
\Phi_\text{middle} = \frac{\max(\Phi_i)}{2}.
\]
To determine the span of the dynamic range, we calculate an interval:
\[
\text{interval} = \Phi_\text{middle} \cdot \frac{\text{range} - 1}{\text{range} + 1},
\]
where \(\text{range} = 8\) is a predefined parameter. The lower and upper bounds for the dynamic range are expressed as:
\[
\Phi_\text{lower} = \Phi_\text{middle} - \text{interval}, \quad 
\Phi_\text{upper} = \Phi_\text{middle} + \text{interval}.
\]
\item 
\textbf{Dynamic Range Clipping.} The radiance values after exposure modeling and noise addition are clipped within the defined bounds:
\[
\Phi_\text{lower} \leq g_i (\Phi_i t_i + n^\text{pre}_i) + n^\text{post}_i \leq \Phi_\text{upper}.
\]

\end{itemize}

\paragraph{Captured Intensity Simulation}
To normalize the captured intensity values to the camera’s range \([0, 1]\), the clipped intensity is processed as:
\[
I^c_{i, \text{captured}}(p^c_i) = \frac{I^c_i(p^c_i) - \min(I^c_i(p^c_i))}{\max(I^c_i(p^c_i)) - \min(I^c_i(p^c_i))}.
\]

\paragraph{Quantization}
Quantization is a critical step in the image formation model, simulating the limited bit depth of real-world cameras by mapping scene radiance values to discrete intensity levels. This process involves scaling, clamping, and rounding intensity values to match the resolution of the target camera system, typically 8 bits. To achieve this, we use a quantization function defined as:
\begin{equation}
\mathrm{quant}(x) = \frac{\operatorname{round}(\operatorname{clip}(x \cdot (2^8 - 1)))}{2^8 - 1}.
\end{equation}
Here, the $\operatorname{clip}(\cdot)$ operation restricts the intensity values to the valid dynamic range, and the $\operatorname{round}(\cdot)$ operation maps the scaled values to the nearest discrete level. This approach ensures that the simulated intensity values align with the physical constraints of stereo cameras while maintaining compatibility with captured image formats.

\paragraph{Straight-Through Estimator (STE) for Backpropagation}
To preserve gradient flow during training, the Straight-Through Estimator (STE) framework is employed for the quantization step. STE approximates the quantization operation as an identity function during the backward pass, effectively bypassing its non-differentiable nature. The gradient of the quantized intensity \(I^\text{quant}_i\) with respect to the scaled intensity \(I^\text{scaled}_i\) is expressed as:
\begin{equation}
\frac{\partial I^\text{quant}_i}{\partial I^\text{scaled}_i} \approx 1.
\end{equation}
This approximation ensures that the quantization operation does not hinder the optimization process, allowing seamless end-to-end training of the model. By combining quantization with STE, the image formation pipeline effectively replicates the behavior of real-world cameras while remaining fully differentiable.

% ###########################################################################

\section{Experimental Prototype}
\label{sec:imaging_system}

\subsection{Device Part List}
Our imaging system consists of a stereo camera, a mobile robot platform, a PC and a 3D LiDAR sensor. The components are selected and configured to ensure synchronized data capture and geometric consistency across diverse environments:
\begin{itemize}
\item \textbf{Stereo Cameras}: Two LUCID Triton 5.4MP cameras (TRI054S-CC) capture 24-bit linear RAW Bayer color images. The cameras are connected via Ethernet and synchronized using the Precise Time Protocol (PTP), achieving sub-millisecond shutter synchronization. For exposure setting at 10 ms with a gain of 1.0, this configuration achieves up to 120 dB of dynamic range in daytime scenes.
\item \textbf{Mobile Platform}: To capture images in diverse real-world environments, we employed the AgileX Ranger-Mini 2.0, a robust four-wheel robot capable of traversing challenging terrains, including urban streets, pedestrian walkways, and indoor environments.
\item \textbf{LiDAR Sensor}: The Ouster OS-1 3D LiDAR sensor provides geometric data with 128 vertical beams, a maximum detection range of 200 meters, and up to 2048 samples per rotation at 20 Hz. The LiDAR’s output resolution reaches \(2048 \times 128\), offering precise depth data. The LiDAR is aligned with the left stereo camera to generate sparse depth maps for the left camera view.
\end{itemize}

\begin{table}[t]
\centering
\begin{tabular}{|c|l|c|l|}
\hline
\textbf{Item \#} & \textbf{Part description} & \textbf{Quantity} & \textbf{Model name} \\ \hline
1 & RGB Camera        & 2 & LUCID Triton TRI054S-CC \\ \hline
2 & Objective lens    & 2 & Edmund Optics \#33-307 \\ \hline
3 & Mobile Platform   & 1 & AgileX Ranger-Mini 2.0 \\ \hline
4 & LiDAR             & 1 & Ouster OS-1 128 \\ \hline
5 & PC                & 1 & ASUS Rog Zephyrus G14 \\ \hline
\end{tabular}
\caption{\textbf{Part list of out imaging system.}}
\label{tab:hardware}
\end{table}

\subsection{Image Acquisition Pipeline}
Our system is designed to capture synchronized stereo and LiDAR data in real-time. The acquisition process is split into two parallel loops:
\begin{enumerate}
\item \textbf{Stereo Image Capture}: The stereo cameras operate at a fixed frame rate of 5 FPS, capturing synchronized frames as 24-bit HDR images saved in \texttt{.npy} format. The cameras are triggered simultaneously at the start of each sequence, ensuring precise temporal alignment.
\item \textbf{LiDAR Data Capture}: The LiDAR sensor scans the environment continuously, sending acknowledgments (ACKs) for each frame. If a corresponding stereo frame is captured within 50 milliseconds of the LiDAR frame, the system associates the two, creating a single synchronized data frame.
\end{enumerate}
This pipeline ensures that stereo intensity data and LiDAR measurements are aligned, enabling robust integration for depth estimation and scene analysis.

\subsection{Calibration details}
\paragraph{Geometric Calibration}
Geometric calibration is performed to align the stereo camera and LiDAR sensor. The calibration parameters include:
\begin{itemize}
\item \textbf{Stereo Cameras}: Intrinsic matrices (focal length, principal point), distortion coefficients, and extrinsic parameters (rotation and translation) are computed using a checkerboard pattern with OpenCV.
\item \textbf{LiDAR-Camera Alignment}: The extrinsic transformation matrix between the LiDAR and the left camera is calculated to project LiDAR points onto the left camera’s image plane, using the camera's intrinsic matrix.
\end{itemize}

\paragraph{Radiometric Calibration}
To ensure consistent intensity measurements across stereo images, the camera settings (exposure and gain) are fixed, and intensity normalization is applied to compensate for sensor sensitivity differences. This step is critical for maintaining accurate depth alignment between the stereo cameras and LiDAR.

\paragraph{Calibration Dataset}
The calibration process uses 50 checkerboard images captured across various distances and angles to optimize the stereo rectification and LiDAR alignment. Reprojection error analysis confirms the geometric accuracy of the calibration parameters.

\section{Datasets}

\subsection{Stereo Real Video Dataset}
\begin{figure}[H]
  \centering
  \includegraphics[width=\linewidth]{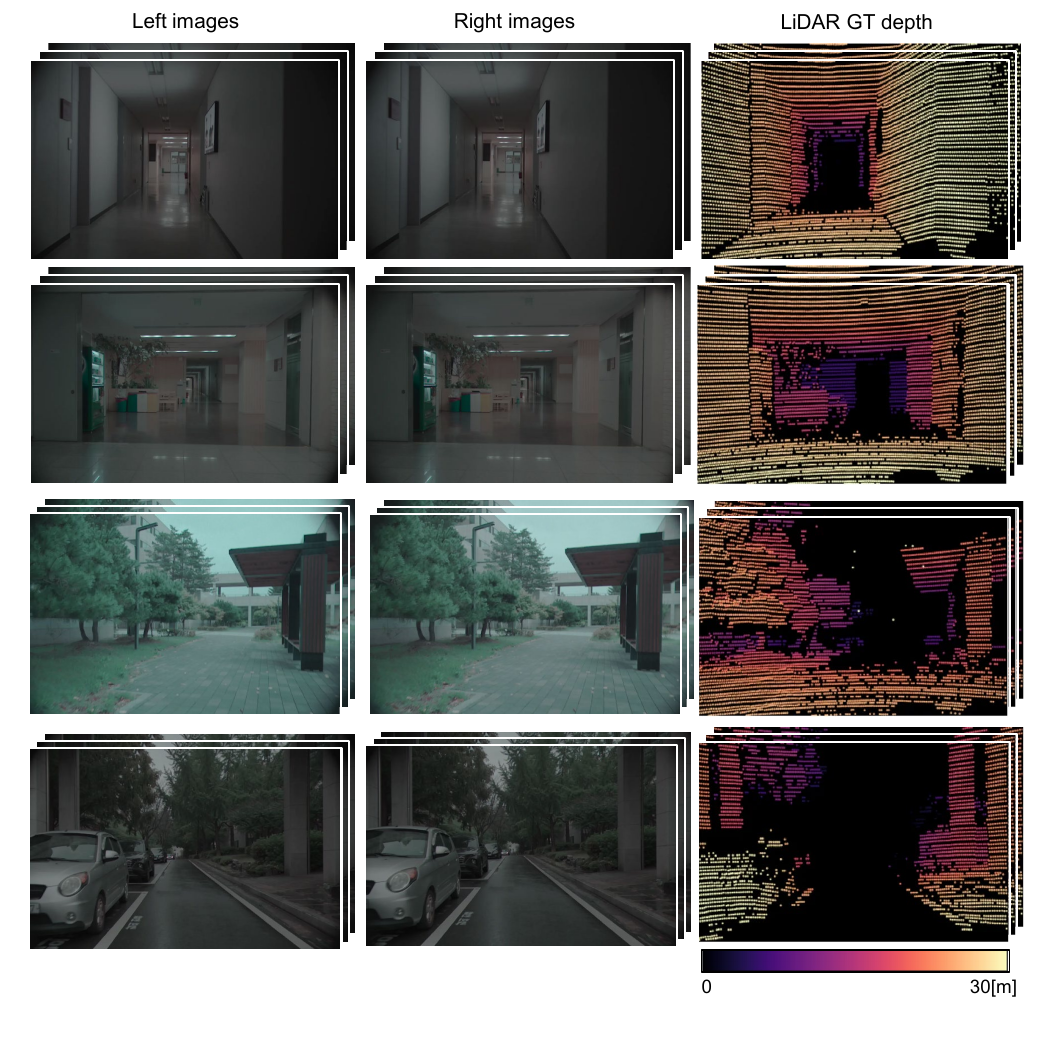}
  \caption{\textbf{Visualization of Real Dataset Samples.} Examples of dual-exposure stereo images and their corresponding LiDAR ground-truth depth maps from the captured real-world dataset. The top two rows represent indoor scenes, while the bottom two rows represent outdoor scenes. The LiDAR GT depth maps demonstrate the variability in point density and accuracy across different environments.}

  \label{fig:Real_dataset_composition}
\end{figure}

The dataset was captured using our stereo camera system described in main paper Section3, equipped with two LUCID Triton 5.4MP cameras for synchronized stereo imaging. Each stereo frame is accompanied by corresponding LiDAR ground truth data captured using an Ouster OS-1 3D LiDAR sensor.
Figure~\ref{fig:Real_dataset_composition} illustrates sample stereo image pairs from the dataset, along with their corresponding ground truth LiDAR points projected onto the left camera view. The stereo images showcase the variety of environments and lighting conditions present in the dataset. The LiDAR ground truth highlights the sparse yet accurate depth information used for evaluation. 

\subsection{Stereo Synthetic Video Dataset}
We use the CARLA driving simulator~\cite{dosovitskiy2017carla} to generate a synthetic video dataset that supports training and testing of dual-exposure stereo depth estimation in diverse automotive scenarios. Our synthetic dataset is specifically configured to capture extreme lighting scenes to simulate real-world dynamic range challenges. To simulate stereo imaging, we configured the CARLA environment with virtual side-by-side mounted RGB-D cameras to capture synchronized stereo image pairs at 1280$\times$384 resolution. Each virtual RGB camera captures full 32-bit stereo images using multi-exposure imaging~\cite{debevec1997hdr}, while the depth camera generates a dense ground truth depth map for each frame. Hereby, the setup generates ground-truth depth maps, ground-truth disparity maps, and stereo calibration data alongside stereo images, enabling the creation of a comprehensive dataset with precise geometry and calibration details consistently across diverse driving scenarios.

\begin{figure*}[t]
  \centering
  \includegraphics[width=\textwidth]{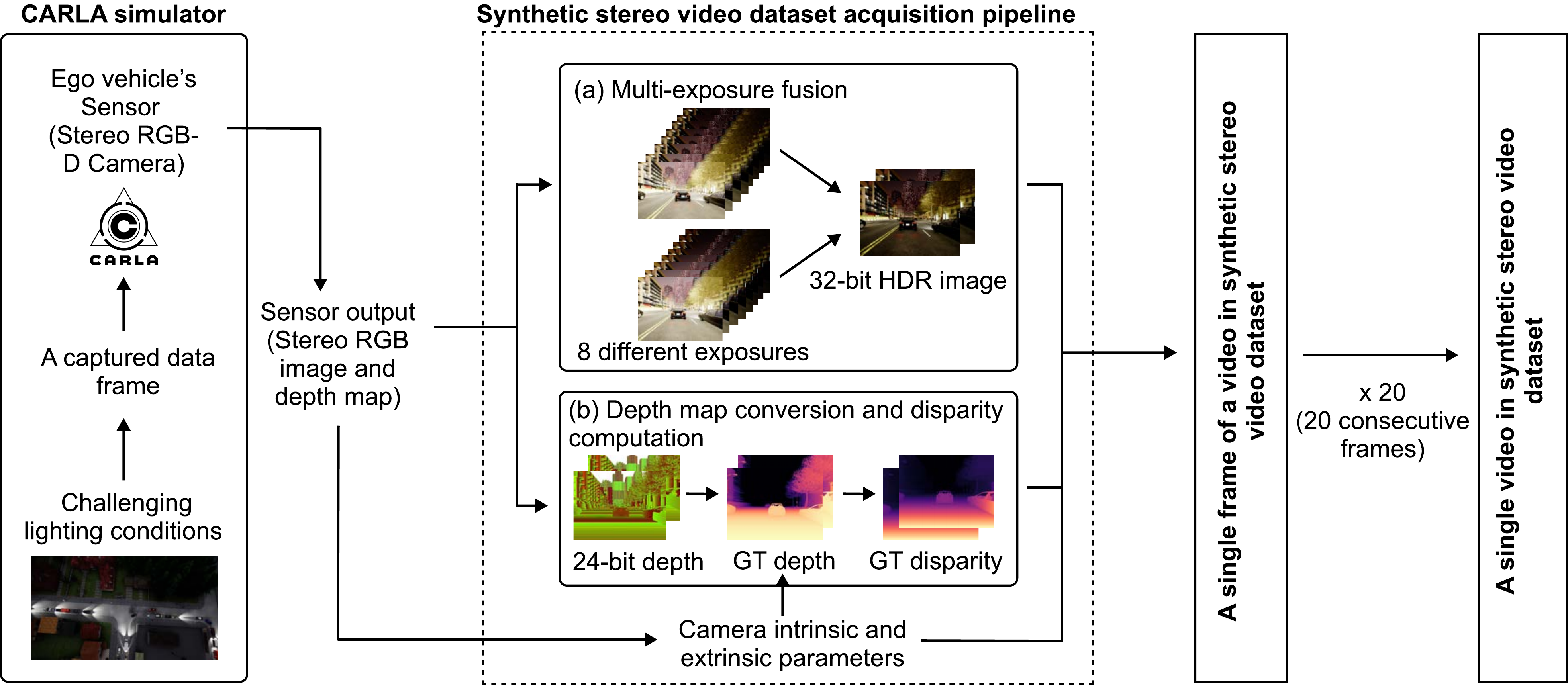}
  \caption{\textbf{Overview of the synthetic stereo video dataset acquisition pipeline.} With specific environmental settings on CARLA simulator to generate adverse lighting conditions, the equipped RGB cameras capture images to reconstruct HDR scenes and the depth cameras create corresponding ground truth disparity maps.}
  \label{fig:acquisition}
\end{figure*}

\paragraph{Dataset Acquisition Pipeline}
Figure~\ref{fig:acquisition} shows an overview of our stereo synthetic video dataset acquisition pipeline. In our CARLA simulator ego-vehicle's capture setup, stereo RGB cameras and paired depth cameras—each with a resolution of 1280$\times$384 pixels, a horizontal field of view of 75 degrees, and a fixed frame rate of 10 FPS—are mounted side by side on the bonnet of the test vehicle, with a baseline of 0.4 m. 
Refer to Table~\ref{tab:sensor_configuration} for the sensor configuration details. Since the CARLA itself does not offer real-time HDR rendering, a primary process is required to reconstruct HDR images from the rendered RGB images. In each time frame, stereo RGB cameras capture images with eight different exposure times $t \in \left\{ \frac{1}{500n} (sec) \mid n \in \{1, 2, \ldots, 8\} \right\}$ while fixed ISO = 200 and aperture size f/1.4 in day time and dusk time.
For night time, on the other hand, exposure times are given as $t \in \left\{ \frac{1}{50n} (sec) \mid n \in \{1, 2, \ldots, 8\} \right\}$ with fixed ISO = 1,600 and aperture size f/1.4. 
Here, daytime is defined as the period when the solar altitude satisfies $\alpha \geq 3^\circ$, and dusk is defined as the period when the solar altitude satisfies $-3^\circ \leq \alpha < 3^\circ$.
Night time is the complement of these periods, corresponding to the range where $\alpha < -3^\circ$. Then, with multi-exposure HDR reconstruction~\cite{debevec1997hdr}, we obtain 32-bit stereo images for each frame of the scenario.
Note that there are no motion artifacts between multi-exposure frames within a single time step of a dynamic automotive scene, as all RGB images are synchronously captured by virtual RGB cameras in the CARLA simulator.
This eliminates the risk of failure in exposure bracketing-based HDR reconstruction for dynamic scenes, which would otherwise require addressing using various de-ghosting approaches~\cite{prabhakar2020towards, pu2020robust, liu2022ghost}. Meanwhile, the depth camera captures the ground truth depth map up to 1,000m. As the CARLA provides depth information with 24-bit floating-point precision encoded across the three channels of the RGB color space, it is decoded to reconstruct the plain depth map in meters. We also compute disparity map from ground-truth depth map using stereo calibration parameters, here by acquiring the ground-truth value for disparity. In specific, given a pair of rectified stereo depth maps with depth $z_s$, focal length $f$ and baseline $B$, the disparity $d$ in the corresponding pixel is calculated using $f\frac{B}{z_s}$. 
As a result, pairs of 32-bit stereo RGB images, depth maps, disparity maps, and stereo calibration parameters (both intrinsic and extrinsic) compose a single frame of a video in the stereo synthetic video dataset, see Table~\ref{tab:dataset_composition}. Additionally, by leveraging CARLA's support for simulating diverse driving environments, both training and testing videos are selectively retrieved from the simulation, introducing abrupt changes in dynamic range and thereby reflecting real-world dynamic range challenges.

\begin{table}[t]
    \centering
    \resizebox{0.8\linewidth}{!}{
    \begin{tabular}{l|c|c|l}
        \toprule[1pt]
        \textbf{Sensor Type} & \textbf{Sensor Count} & \textbf{Output Shape} & \textbf{Configuration}\\ \midrule
        RGB camera & 
        8 & 
        $\mathbb{R}^{3 \times 384 \times 1280}$ & 
        Left, ISO = 200(Day, Dusk) / 1600(Night), f/1.4, FOV = $75^\circ$ \\

        RGB camera & 
        8 & 
        $\mathbb{R}^{3 \times 384 \times 1280}$ & 
        Right, ISO = 200(Day, Dusk) / 1600(Night), f/1.4, FOV = $75^\circ$\\

        Depth sensor & 
        1 & 
        $\mathbb{R}^{1 \times 384 \times 1280}$ & 
        Left, FOV = $75^\circ$\\

        Depth sensor & 
        1 & 
        $\mathbb{R}^{1 \times 384 \times 1280}$ & 
        Right, FOV = $75^\circ$\\ 
        \bottomrule[1pt]
    \end{tabular}
    }
    \caption{\textbf{List of sensors used for CARLA simulator ego-vehicle's capture setup.} Here, ISO is configured based on the temporal condition. Four categories of sensors are mounted at \( x = 2.5 \, \text{m} \), \( y = \pm 0.2 \, \text{m} \), \( z = 1.4 \, \text{m} \) with respect to the ego-vehicle’s centroid. Note that the given coordinates follow the left-handed coordinate system in Unreal Engine 4. }
    \label{tab:sensor_configuration}
    
\end{table}

\begin{table}[t]
    \centering
    \resizebox{0.7\linewidth}{!}{
    \begin{tabular}{l|c|l}
        \toprule[1pt]
        \textbf{Modality} & \textbf{Shape} & \textbf{Description} \\ \midrule
        HDR image & 
        $\mathbb{R}^{3 \times 384 \times 1280}$ & 
        A pair of left and right, 32-bit float \\

        Depth map & 
        $\mathbb{R}^{1 \times 384 \times 1280}$ & 
        A pair of left and right, up to 1,000m \\

        Disparity map& 
        $\mathbb{R}^{1 \times 384 \times 1280}$ & 
        A pair of left and right, computed from depth map\\

        Intrinsic camera parameters & 
        $\mathbb{R}^{3 \times 3}$ & 
        Shared between the left and right views \\

        Extrinsic camera parameters & 
        $\mathbb{R}^{4 \times 4}$ & 
        A pair of left and right \\ 
        \bottomrule[1pt]

    \end{tabular}
    }
    \caption{\textbf{Dataset composition for a single frame in a stereo synthetic video dataset.} Each modality, its dimensions, and additional details are outlined.}
    \label{tab:dataset_composition}
\end{table}

\begin{figure*}[t]
  \centering
  \includegraphics[width=\textwidth]
  {figures/synthetic_dataset_thumbnails.pdf}
  \vspace{-6mm} 
  \caption{\textbf{Synthetic stereo video datasets} (a) Sample tone-mapped thumbnails with diverse driving environments. (b) Dataset scene label statistics for two versions of datasets.}
  \vspace{-3mm}
  \label{fig:dataset_thumbnails}
\end{figure*}

\paragraph{Dataset Details and Statistics}
Our synthetic dataset comprises 1,000 training videos and 200 testing videos, each designed to introduce dynamic range challenges across various driving conditions. Training scenarios comprise 20 consecutive stereo frames, and test scenarios contain 100 consecutive stereo frames. Scenarios represent a wide range of lighting conditions (day, dusk, and night), with distributions of approximately 50\% at night, 30\% during the day, and 20\% at dusk. The driving locations include urban and suburban areas, rural areas, and highways.
Each video presents challenging lighting conditions induced by various environmental factors, categorized into four major types: the vehicle's headlights at night, intense reflections from highly reflective surfaces (such as ponds), intense natural lighting, and light passing through tunnels. 
Figure~\ref{fig:dataset_thumbnails} shows the dataset thumbnails and scene statistics of the synthetic dataset, which includes diverse driving environments.

\begin{figure*}[t]
  \centering
  \includegraphics[width=0.6\textwidth]{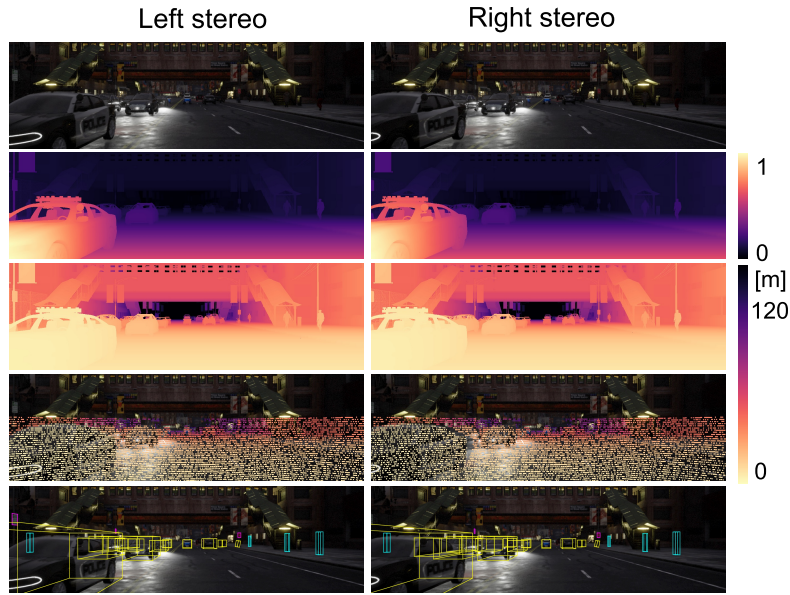}
  \caption{\textbf{Extended synthetic stereo video datasets.} The bottom two rows depict LiDAR point clouds and 3D bounding boxes projected onto the tone-mapped stereo HDR images. To ensure consistency across different modalities, both the depth and LiDAR maps share the same color bar, while the disparity map is shown after normalization.}
  \label{fig:synthetic_dataset_extended}
\end{figure*}

\paragraph{Extended Dataset for 3D Object Detection}
To extend the baseline dataset for the vision task of object detection, we introduce an additional dataset that facilitates both 2D and 3D object detection. Extended dataset consists of 750 driving videos for training and 150 videos for testing, both adhering to the same specifications and scene diversities as the baseline dataset, with the addition of two new modalities: (1) LiDAR point clouds, (2) per-frame object detection data annotations. The virtual LiDAR system is configured to replicate the characteristics of a Velodyne HDL-64E (64 channels, 10Hz revolution frequency, from $-24.8^\circ$ to $+2^\circ$ vertical field of view and 120m maximum range) and mounted along the optical axis of the left RGB-D camera.
The cameras' exposures are triggered only when the LiDAR has completed its rotation and is aligned with the optical axis of the left camera, ensuring precise cross-modality alignment between the LiDAR and the stereo RGB-D cameras.
For each object within the left camera's field of view, we automatically annotate it using 3D bounding boxes in a format simplified from the KITTI~\cite{Geiger2013IJRR} object detection labels. The annotations include fields for class names, truncation, occlusion state, and bounding box coordinates, all represented in the reference camera's coordinate system.
Specifically, we provide annotations for three object classes: `Vehicle', `Pedestrian', `Traffic Signal'. Figure~\ref{fig:synthetic_dataset_extended} presents the composition of the extended dataset, highlighting the time-varying lighting conditions that contribute to challenging lighting conditions for both depth estimation and object detection.

% ###################################################################

\section{Dual-Exposure Depth Estimation}
\subsection{Network Architecture}

Our dual-exposure depth estimation model extends the RAFT-Stereo framework~\cite{lipson2021raft} by incorporating modules for dual-exposure feature fusion and inter-frame motion compensation. These additions enable the network to effectively utilize exposure-specific features from dual-exposure stereo inputs, enhancing disparity estimation under high dynamic range conditions.

\paragraph{Network Overview}
The architecture consists of three primary stages: 
1) Optical Flow Estimation, 
2) Dual-Exposure Feature Fusion, and 
3) Stereo Depth Estimation. 
These components are seamlessly integrated into the RAFT-Stereo backbone. While the backbone's original disparity estimation modules remain unchanged, modifications were made to handle dual-exposure inputs and inter-frame alignment:
% \paragraph{Key Modifications and Additions}
\begin{itemize}
    \item \textbf{Optical Flow Estimation:} 
    We introduce a pretrained optical flow network~\cite{morimitsu2024rapidflow} to estimate motion between consecutive frames for both left and right stereo views. The optical flow enables spatial alignment of the second-frame features to the first-frame features, addressing temporal motion.
    
    \item \textbf{Dual-Exposure Feature Fusion:} 
    A feature fusion module combines aligned features from dual-exposure stereo frames. This module uses intensity-based weight maps to ensure that well-exposed details from both bright and dark regions are effectively preserved. The fusion is applied at multiple scales to enhance robustness.
    
    \item \textbf{Stereo Disparity Estimation:} 
    The fused features are passed through the RAFT-Stereo backbone to construct correlation volumes and refine disparity predictions. While the correlation computation and update block follow the original RAFT-Stereo design, they now operate on fused feature maps containing dual-exposure information.
\end{itemize}

\paragraph{Summary of Modified Layers}
Table~\ref{tab:network_architecture} summarizes the layers and modules where significant modifications were made. Components such as the correlation volume and update block are inherited directly from the RAFT-Stereo framework and are not described in detail here.

\begin{table}[h]
    \centering
    \resizebox{\linewidth}{!}{
    \begin{tabular}{l|c|c|c}
        \toprule[1pt]
        \textbf{Module} & \textbf{Input Size} & \textbf{Output Size} & \textbf{Description} \\ \midrule
        Optical Flow Network & $[B, 3, H, W]$ & $[B, 2, H, W]$ & Estimates motion for temporal alignment. \\ 
        Warping Function & $[B, 256, H/4, W/4]$ & $[B, 256, H/4, W/4]$ & Aligns second-frame features using optical flow. \\ 
        Dual-Exposure Fusion & $[B, 256, H/4, W/4]$ & $[B, 256, H/4, W/4]$ & Combines features from dual-exposure frames using intensity-based weights. \\ 
        \bottomrule[1pt]
    \end{tabular}}
    \caption{\textbf{Modified Modules in the Proposed Network.} The table summarizes the key components added to the RAFT-Stereo backbone for dual-exposure depth estimation. Input and output sizes are for a batch size of $B$ and image resolution $H \times W$.}
    \label{tab:network_architecture}
\end{table}

\paragraph{Integration with RAFT-Stereo Backbone}
The proposed modifications are integrated into the RAFT-Stereo backbone while retaining its core functionality. Optical flow is computed between consecutive stereo frames and used to warp second-frame features to the first frame. These warped features are then fused with first-frame features using the dual-exposure feature fusion module. The fused features are passed through the correlation volume computation and update block to estimate the disparity map. This integration ensures that dual-exposure information is effectively utilized while maintaining the robustness of the original RAFT-Stereo design.

\subsection{Training Details}

\paragraph{Data Augmentation and Exposure Simulation}
To simulate dual-exposure stereo inputs, we generated random exposure pairs for each training batch using controlled randomization. The exposure values \(e_1\) and \(e_2\) for the two frames were generated as:
\[
e_2 = e_1 \cdot \text{rand}(\text{min\_gap}, \text{max\_gap}),
\]
where \(e_1, e_2 \in [2^{-2}, 2^2]\) and the gap \(\text{rand}(\text{min\_gap}, \text{max\_gap})\) was sampled uniformly between 0.5 and 3.0. This exposure simulation ensures the model is trained across diverse lighting conditions, reflecting real-world variations in dynamic range.

\paragraph{Loss Function}
For training, we employed the sequence loss function adopted from the original RAFT-Stereo framework~\cite{lipson2021raft}. This loss progressively refines disparity predictions over \(N = 32\) iterations, with a decay factor \(\gamma = 0.9\). The sequence loss is defined as:
\[
\mathcal{L}_{\text{seq}} = \sum_{i=1}^{N} \gamma^{\frac{15}{N-1}(N-i-1)} \cdot \| \hat{d}_i - d_{\text{gt}} \|_1,
\]
where \(\hat{d}_i\) represents the predicted disparity at iteration \(i\), and \(d_{\text{gt}}\) is the ground truth disparity. A validity mask filters out invalid regions and restricts the loss to valid pixels with a maximum disparity threshold of 700. This ensures effective training of disparity refinement while avoiding the impact of large outliers.

\paragraph{Training Configuration}
The training was conducted on four NVIDIA RTX 3090 GPUs with a batch size of 4. The CARLA synthetic dataset was used to simulate extreme lighting scenarios, providing diverse and challenging conditions for training. To preserve the robustness of the pretrained RAFT-Stereo model on real-world datasets, only the GRU update block was fine-tuned during training. All other layers were frozen to retain their existing weights. This targeted fine-tuning strategy ensured that the network specialized in feature fusion and disparity refinement for dual-exposure inputs, without degrading its performance on real datasets. 

\section{Additional Results}
\subsection{Additional Evaluation on Real Dataset}

\begin{figure}[H]
  \centering
  \includegraphics[width=\linewidth]{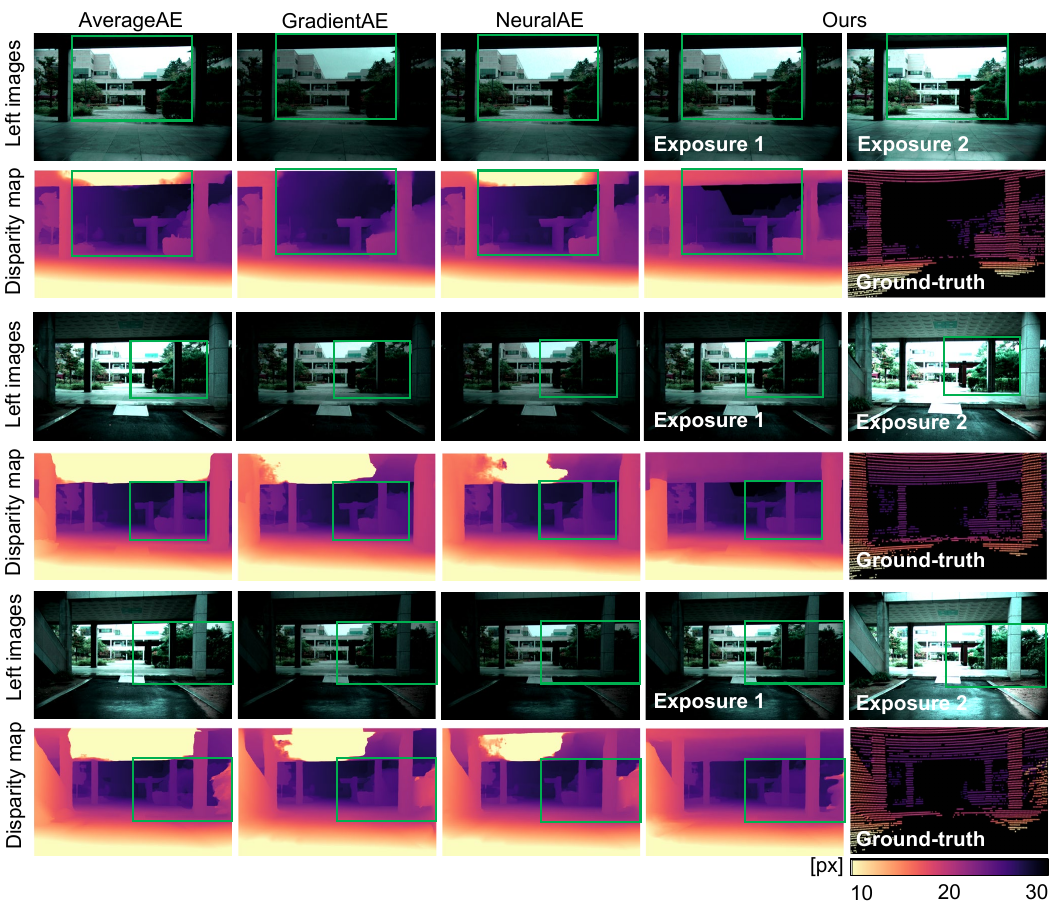}
  \caption{\textbf{Disparity-estimation results using our ADEC compared with other AEC methods in outdoor scene} Our ADEC method outperforms the other AEC methods for subsequent extended-DR depth estimation : AverageAE~\cite{cameraProduct}, GradientAE~\cite{shim2018gradient}, NeuralAE~\cite{onzon2021neural}}
  \label{fig:Supp_real_add_comparison}
\end{figure}

\begin{figure}[H]
  \centering
  \includegraphics[width=\linewidth]{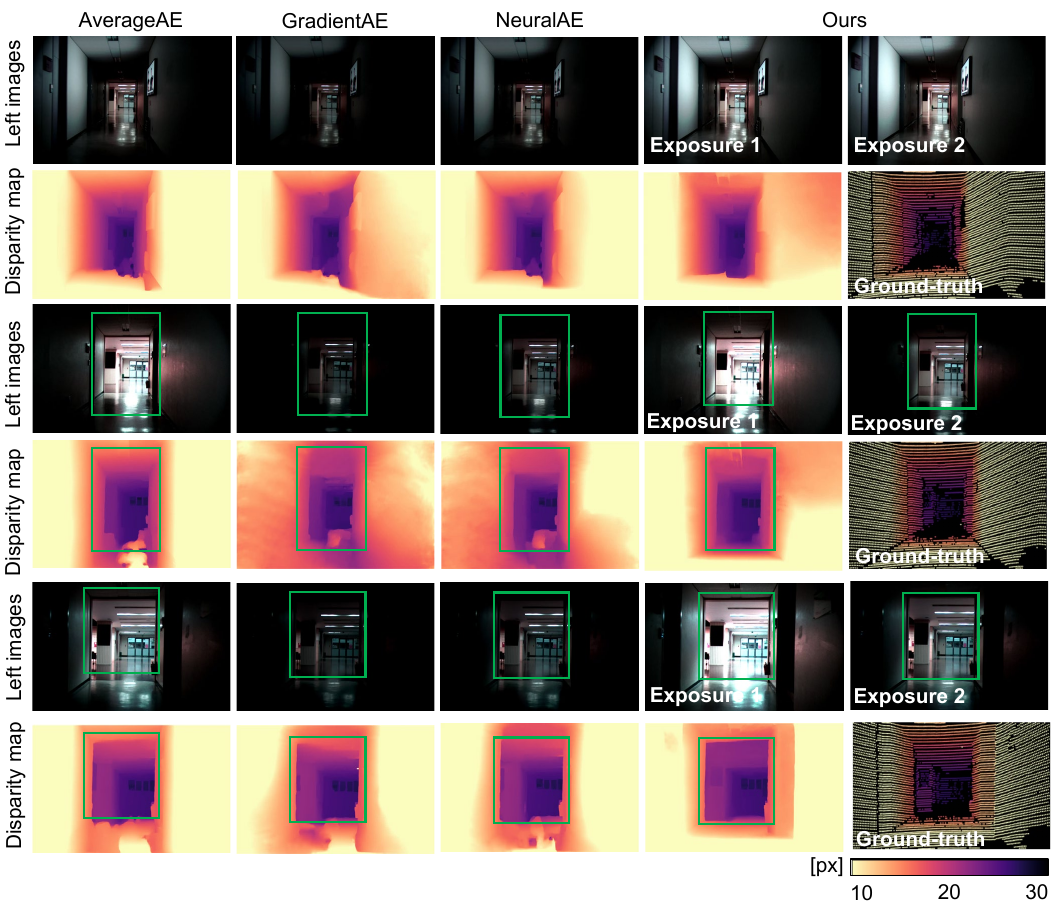}
  \caption{\textbf{Disparity-estimation results using our ADEC compared with other AEC methods in indoor scene} Our ADEC method outperforms the other AEC methods for subsequent extended-DR depth estimation : AverageAE~\cite{cameraProduct}, GradientAE~\cite{shim2018gradient}, NeuralAE~\cite{onzon2021neural}}
  \label{fig:Supp_real_add_comparison2}
\end{figure}

To further validate our method, we conducted evaluations on real-world scenarios featuring dynamic lighting changes. These scenarios include both indoor and outdoor environments, emphasizing the robustness of our approach under challenging illumination conditions. The evaluation comprises four distinct scenarios, consisting of approximately 1000 frames in total. Figure~\ref{fig:Supp_real_add_comparison} visualizes the results of outdoor scenes, with each row representing a consecutive frame in temporal order, showcasing the effectiveness of our method in handling dynamic lighting changes across time. Similarly, Figure~\ref{fig:Supp_real_add_comparison2} demonstrates the results for indoor scenes, where the images also follow a temporal sequence. 

\begin{table}[h]
    \centering
    \resizebox{0.7\linewidth}{!}{
    \begin{tabular}{l|c|c|c|c}
        \toprule[1pt]
        \textbf{Method} & \textbf{AverageAE}~\cite{cameraProduct} & \textbf{GradientAE}~\cite{shim2018gradient} & \textbf{NeuralAE}~\cite{onzon2021neural} & \textbf{ADEC (ours)} \\ \midrule
        MAE [m]\(\downarrow\) & 2.6142 & 4.1859 & \underline{2.2869} & \textbf{2.0251} \\ 
        \bottomrule[1pt]
    \end{tabular}
    }
    \caption{\textbf{Comparison of MAE across methods.} The table highlights the performance of different methods, showing that our approach (ADEC) achieves the lowest MAE compared to other baselines.}
    \label{tab:mae_comparison}
\end{table}

\subsection{Additional Evaluation on Synthetic Dataset}

\begin{figure}[H]
  \centering
  \includegraphics[width=\linewidth]{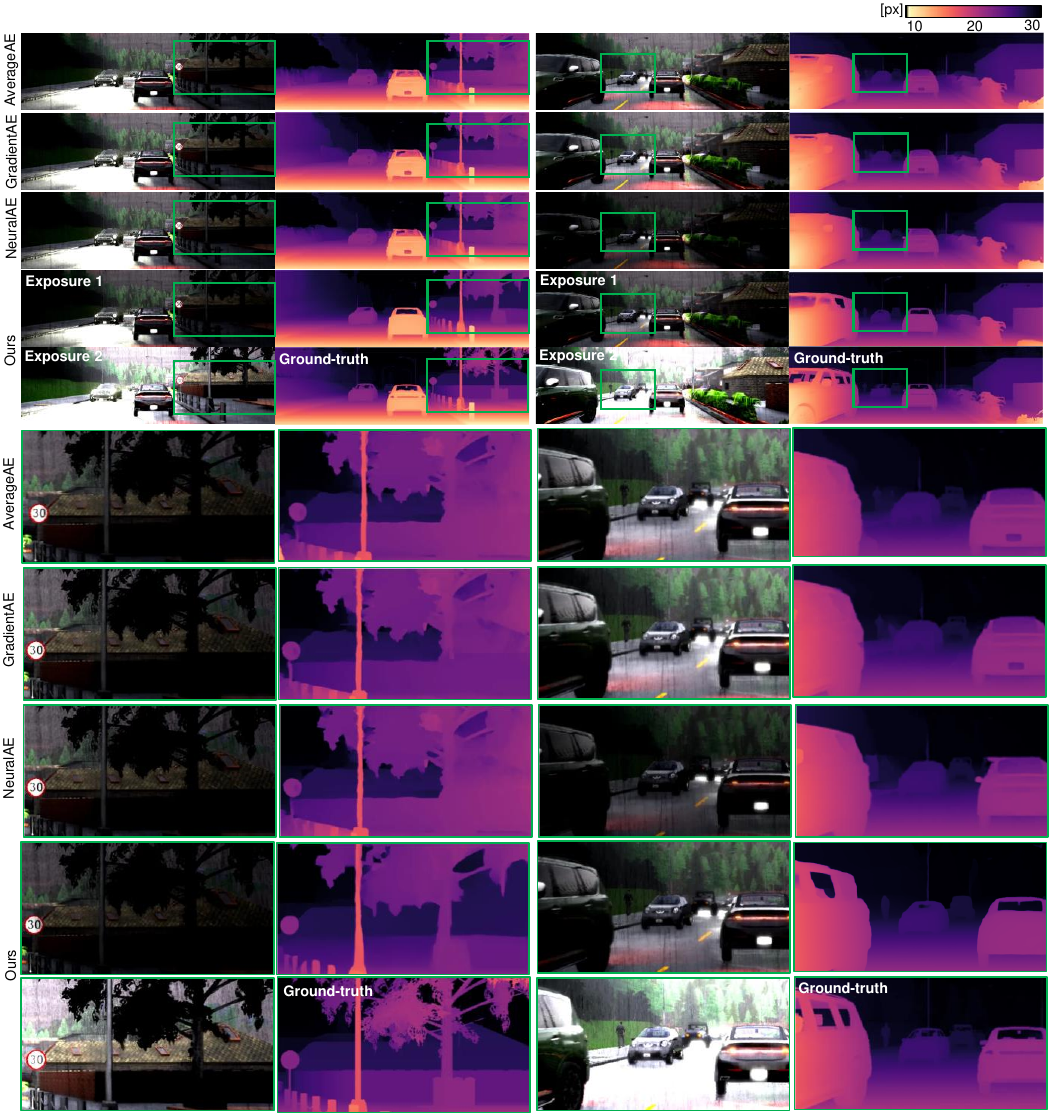}
  \caption{\textbf{Disparity-estimation results using our ADEC compared with other AEC methods} Our ADEC method outperforms the other AEC methods for subsequent extended-DR depth estimation : AverageAE~\cite{cameraProduct}, GradientAE~\cite{shim2018gradient}, NeuralAE~\cite{onzon2021neural}.}
  \label{fig:Supp_synthetic_add_comparison}
\end{figure}

We conducted a evaluation of our method on the CARLA synthetic dataset to further demonstrate its robustness under various exposure and lighting conditions. The comparison includes other single exposure control methods : AverageAE~\cite{cameraProduct}, GradientAE~\cite{shim2018gradient}, and NeuralAE~\cite{onzon2021neural} finetuned using the original RAFT-Stereo framework on the our CARLA synthetic dataset. This ensures a fair comparison between our dual-exposure control approach and existing single-exposure control methods. Figure~\ref{fig:Supp_synthetic_add_comparison} illustrates qualitative results comparing disparity maps generated by each method. The dataset includes diverse scenarios, such as high-contrast outdoor environments and challenging low-light conditions. For each method, we show the left image input, the predicted disparity map, and the corresponding ground-truth disparity.

\subsection{Additional Ablation Experiments}
\begin{figure}[H]
  \centering
  \includegraphics[width=\linewidth]{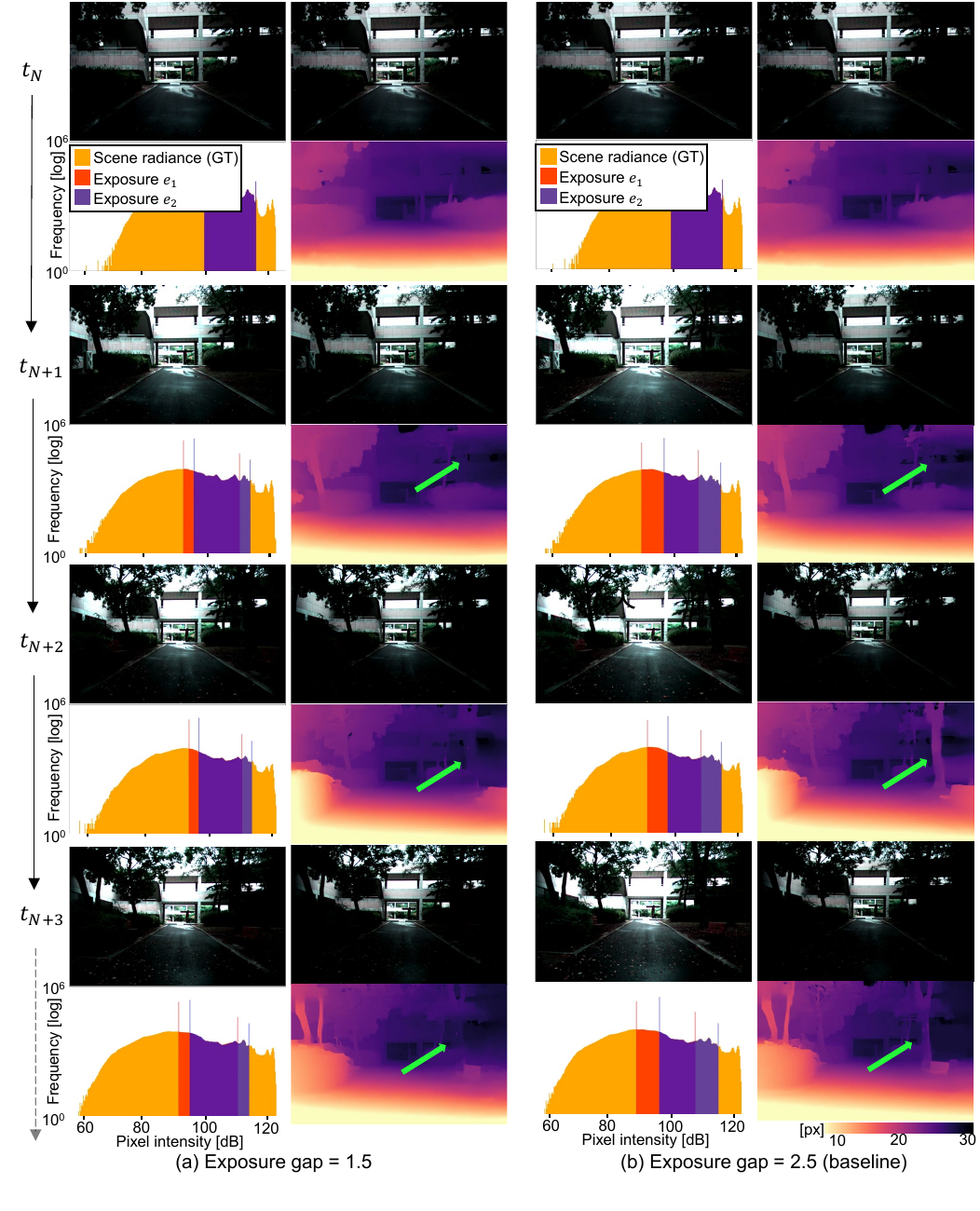}
  \caption{\textbf{Impact of exposure gap settings on disparity estimation.} This figure illustrates the effect of varying the exposure gap during dual-exposure control. The baseline model (exposure gap = 2.5) captures sufficient details over time, whereas the ablation model (exposure gap = 1.5) struggles to widen the exposure gap further, resulting in insufficient detail capture in challenging lighting conditions. Each time step showcases the dual-exposure stereo images, pixel intensity histograms, and disparity maps.}

  \label{fig:Ablation_Exposure_gap_setting}
\end{figure}

\begin{figure}[H]
  \centering
  \includegraphics[width=\linewidth]{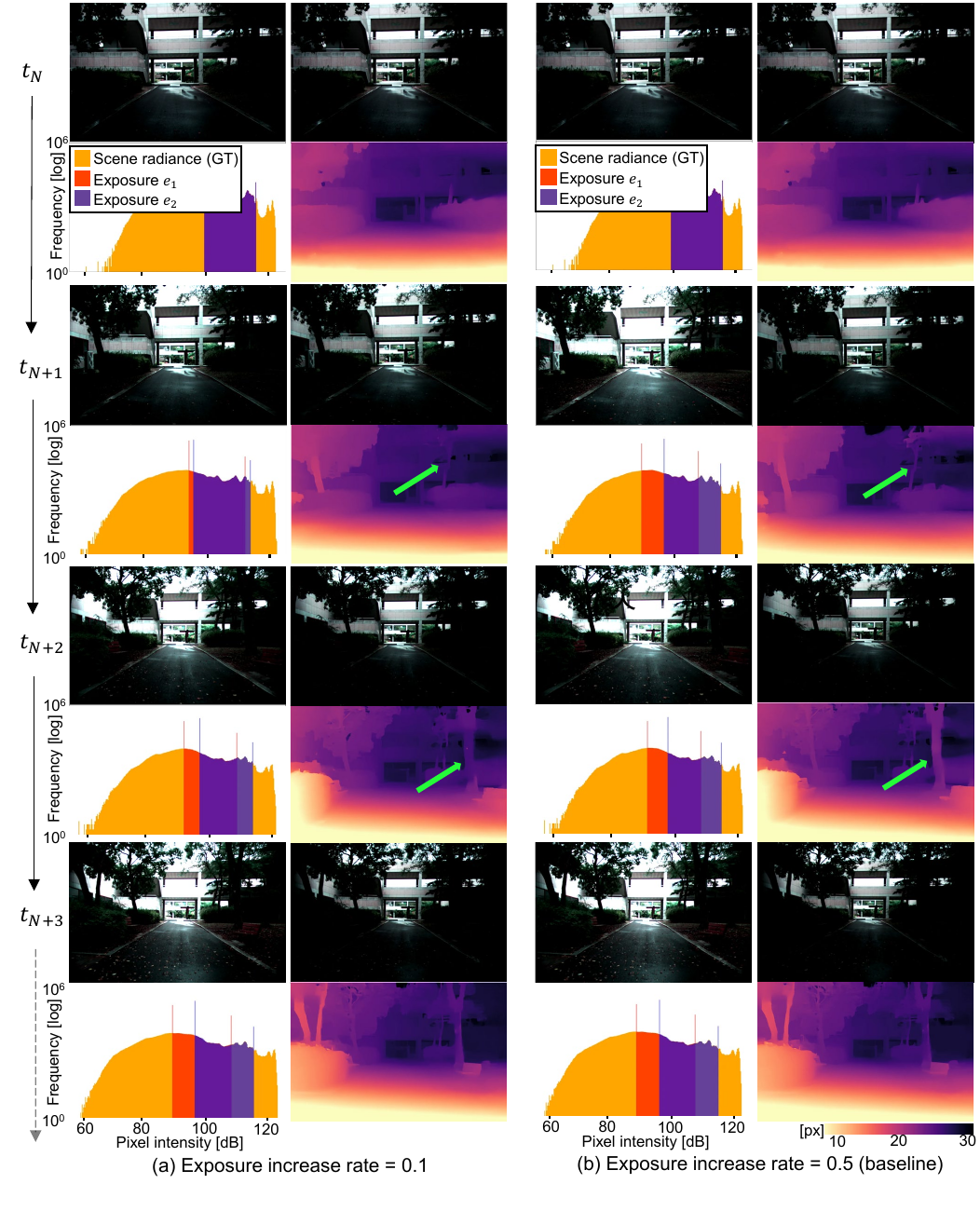}
  \caption{\textbf{Impact of exposure increase rate on disparity estimation.} This figure demonstrates the effect of modifying the exposure increase rate during dual-exposure control. The baseline model, with its default increase rate, quickly expands the exposure gap in the initial time steps, enabling effective detail capture. In contrast, the ablation model, with a reduced increase rate, shows slower gap expansion, leading to less effective detail capture in the early time steps. Each time step visualizes the dual-exposure stereo images, pixel intensity histograms, and disparity maps.}

  \label{fig:Ablation_Exposure_increase_rate}
\end{figure}

\begin{figure}[H]
  \centering
  \includegraphics[width=\linewidth]{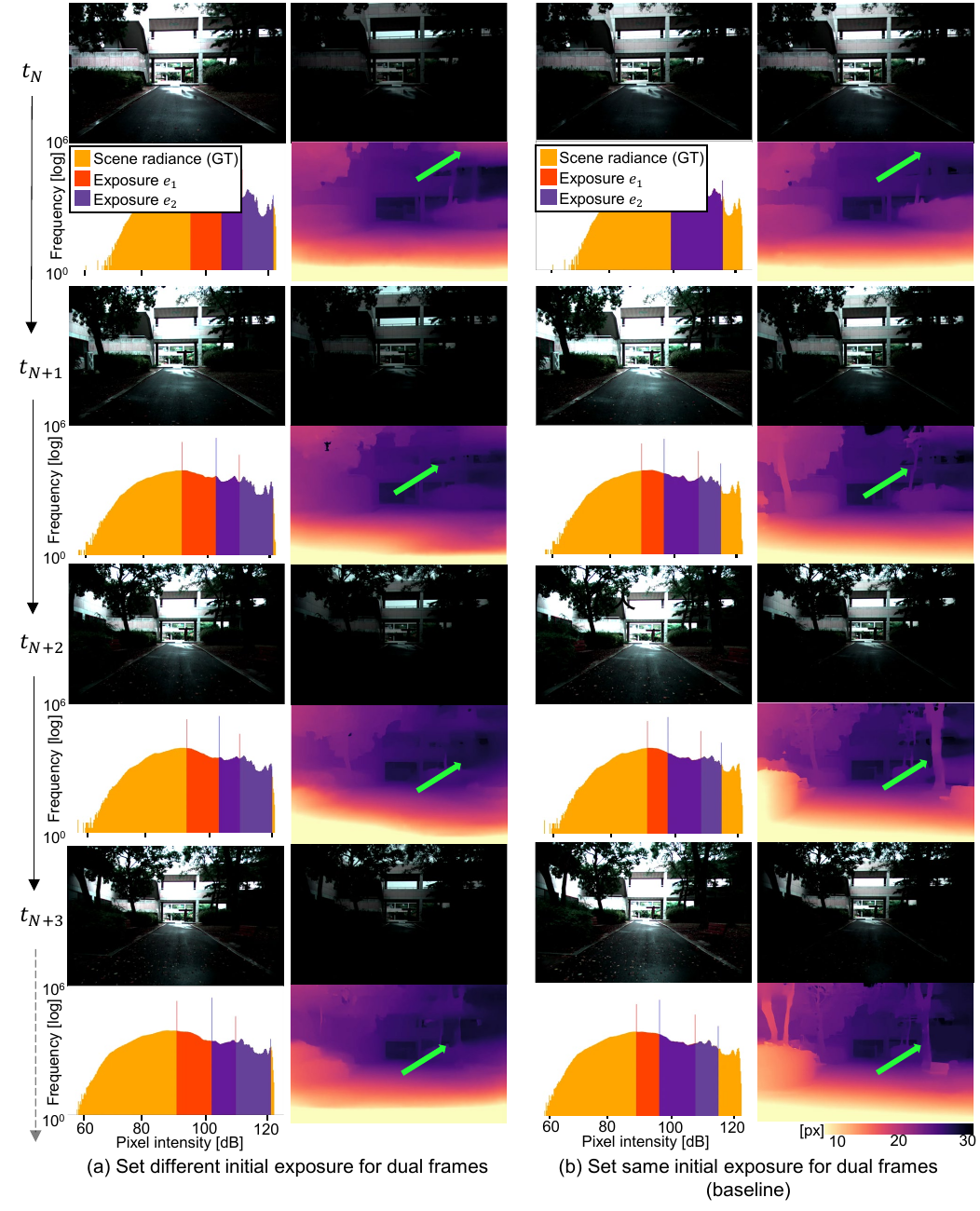}
  \caption{\textbf{Impact of initial exposure settings on disparity estimation.} This figure compares baseline and ablation models with different initial exposures. The baseline model uses equal exposures, ensuring consistent detail capture over time. The ablation model starts with unequal exposures, capturing more detail initially but losing balance in later time steps.}

  \label{fig:Ablation_Initial_exposure_setting}
\end{figure}

We conducted additional ablation studies focusing on the exposure control module to evaluate its impact on performance. The results are presented both quantitatively and qualitatively through Table~\ref{tab:ablation_exposure} and Figures~\ref{fig:Ablation_Exposure_gap_setting}, ~\ref{fig:Ablation_Exposure_increase_rate}, and ~\ref{fig:Ablation_Initial_exposure_setting}. Each figure visualizes the ablation results by comparing the baseline and ablation models across time steps. For each time step, the visualizations include dual-exposure stereo images, pixel intensity histograms, and disparity maps.

\paragraph{Exposure Gap Configuration}
In Figure~\ref{fig:Ablation_Exposure_gap_setting}, we evaluate the effect of different exposure gap configurations. While the baseline model gradually increases the exposure gap, the ablation model fails to widen the gap significantly after a certain point. This results in difficulty capturing sufficient details, particularly in high-contrast regions, compared to the baseline model.

\paragraph{Exposure Increase Rate}
In Figure~\ref{fig:Ablation_Exposure_increase_rate} demonstrates the impact of modifying the scaling factor for determining the next exposure value, referred to as the exposure increase rate. Compared to the baseline model, the ablation model does not achieve a sufficiently large exposure gap in the initial time steps. As a result, the baseline model captures more details in critical regions at earlier time steps, while the ablation model struggles to do so.

\paragraph{Initial Exposure Values}

In Figure~\ref{fig:Ablation_Initial_exposure_setting}, we analyze the effect of setting different initial exposure values for dual-exposure frames. In the baseline model, both exposures start at the same value, while in the ablation model, one frame starts with a higher exposure and the other with a lower one. Although the ablation model benefits from a pre-established exposure gap in the first time step, the baseline model eventually outperforms it by securing more consistent details as the time steps progress.

These results illustrate how variations in exposure gap configuration, exposure increase rate, and initial exposure settings influence the ability of the model to capture and preserve sufficient detail across dynamic scenes. The figures highlight the importance of a well-balanced exposure control strategy for robust disparity estimation.

\begin{table}[t]
    \centering
    \resizebox{0.9\columnwidth}{!}{
        \begin{tabular}{c|c|c|c|c}
        \toprule[1pt]
        \textbf{Experiment} & \textbf{Exposure Gap} & \textbf{Exposure Increase Rate} & \textbf{Initial Exposure Values} & \textbf{Disparity MAE [px]$\downarrow$} \\
        \midrule
        Baseline            & 2.5               & Baseline                       & Equal                          & \textbf{2.7452} \\ % Baseline result
        Ablation 1          & 1.5               & Baseline                       & Equal                          & \underline{2.8759} \\ % Limited Exposure Gap
        Ablation 2          & 2.5               & Reduced                        & Equal                          & 2.7634 \\ % Reduced Exposure Increase Rate
        Ablation 3          & 2.5               & Baseline                       & Unequal                        & 3.2522 \\ % Different Initial Exposure Values
        \bottomrule[1pt]
        \end{tabular}
    }
    \caption{\textbf{Ablation study on exposure control parameters.} The table presents the disparity MAE for different ablation settings, focusing on exposure gap, exposure increase rate, and initial exposure values. The baseline uses an exposure gap of 2.5, baseline increase rate, and equal initial exposure values, achieving the lowest MAE.}
    \label{tab:ablation_exposure}
\end{table}

% \subsubsection{Exposure gap setting}

% \subsubsection{Exposure increase rate}

% \subsubsection{Without weight based feature fusion}
% \todo{XX}

% \subsubsection{Without motion adjustment}
% \todo{XX}

\section{Additional Discussion}
\subsection{Motion blur in dataset acquisition}
While our method demonstrates significant improvements in disparity estimation under challenging lighting conditions, it is not without limitations. One key challenge arises during dataset acquisition, particularly when the stereo cameras are mounted on a mobile robot. Despite careful synchronization of the stereo cameras, as described in Section~\ref{sec:imaging_system}, motion blur can occur in consecutive frames if the mobile robot experiences sharp rotations or vibrations during movement. This motion blur, even in a single frame, can adversely affect our dual-exposure disparity estimation pipeline.

\begin{figure}[H]
    \centering
    \includegraphics[width=\linewidth]{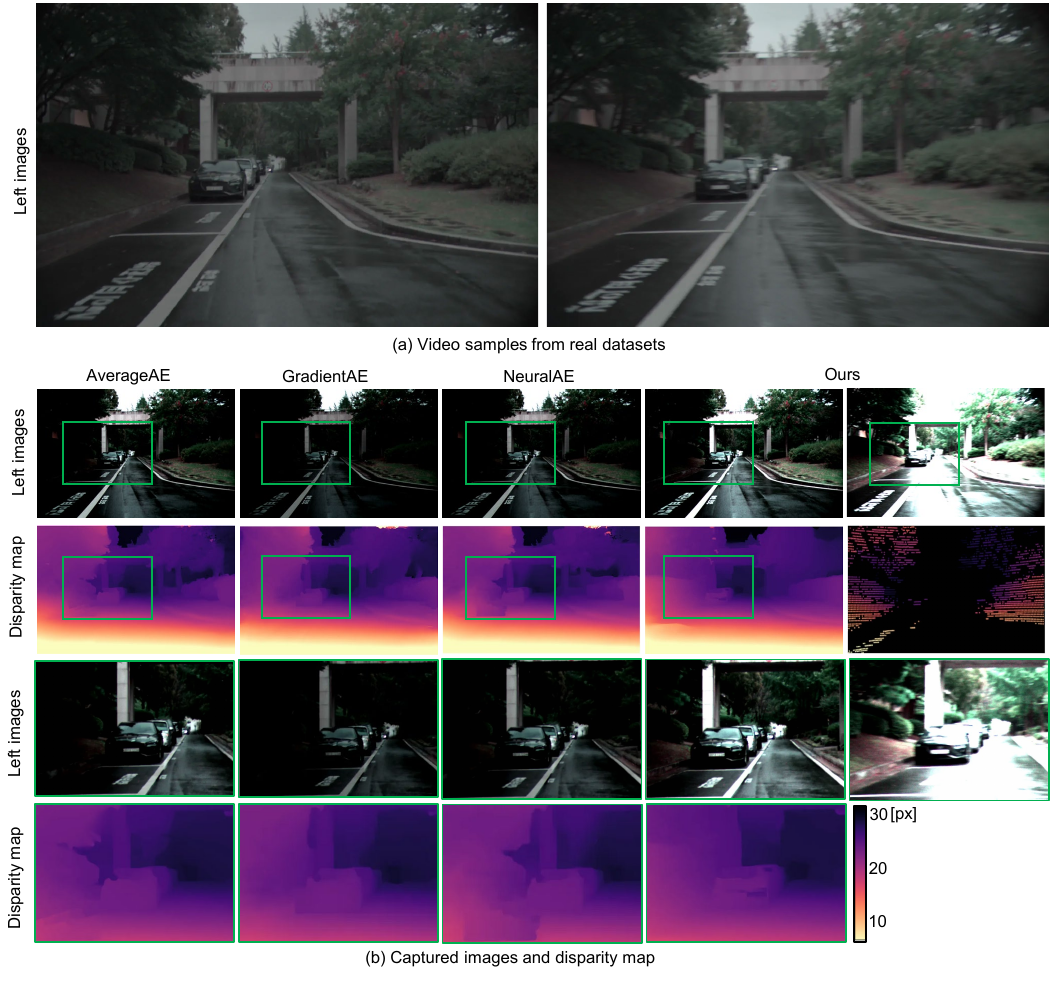}
    \caption{\textbf{Impact of Motion Blur on Disparity Estimation.} (a) An example of motion blur in one frame due to robot movement during dataset acquisition. (b) Disparity maps generated by different methods for the same scene, showing the sensitivity of our method to motion blur.}
    \label{fig:motion_blur}
\end{figure}

Figure~\ref{fig:motion_blur} illustrates an example of this limitation. (a) shows a sample captured from our dataset, where one of the frames exhibits motion blur due to the robot's movement. (b) compares disparity maps generated by different methods for this scene. The results indicate that our method is particularly sensitive to motion blur, as it relies on the effective fusion of details from consecutive frames. The blurred frame reduces the accuracy of feature alignment and fusion, ultimately impacting the disparity estimation. 

To address the limitations posed by motion blur, several strategies can be explored. First, robust feature extraction techniques could be employed to reduce the sensitivity to motion blur. This could include pre-processing steps such as deblurring algorithms or using motion-compensated encoders to improve the quality of extracted features. Second, frames affected by severe motion blur can be automatically detected and excluded from training or evaluation using motion blur detection algorithms that analyze temporal or spatial gradients. Lastly, employing higher frame rate cameras during dataset acquisition could significantly reduce motion blur by capturing images at shorter time intervals, thereby improving the alignment and fusion of stereo features in our pipeline. These solutions offer promising directions to enhance the robustness of our method against motion blur while maintaining its effectiveness in challenging scenarios. Future work will focus on implementing these strategies to further enhance the robustness of our method in dynamic real-world scenarios.

\subsection{Challenges with LiDAR points in outdoor scenarios}
\begin{figure}[H]
    \centering
    \includegraphics[width=\linewidth]{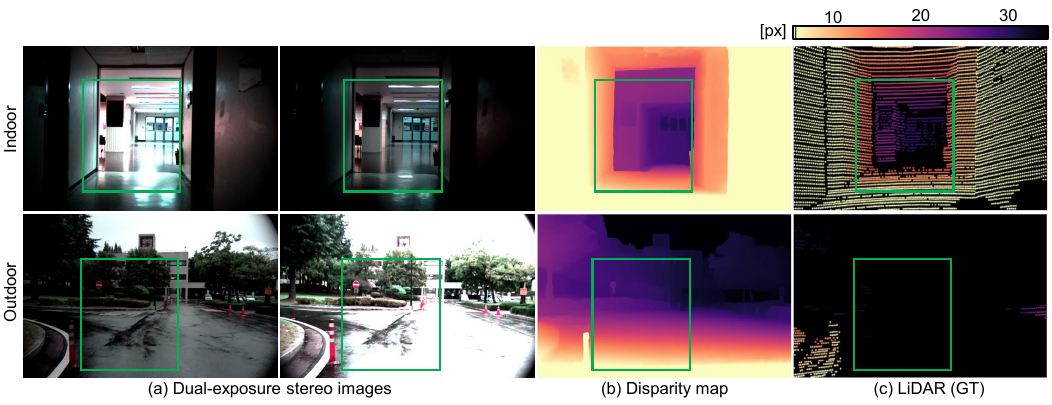}
    \caption{\textbf{Challenges with LiDAR Points in Indoor and Outdoor Scenarios.} 
    (a) Dual-exposure stereo images for indoor and outdoor scenes. (b) Disparity maps generated by our method, showing accurate reconstruction for indoor and outdoor scenes (c) LiDAR ground-truth points, illustrating the variation in point density between indoor and outdoor scenes, particularly on wet ground in the outdoor scenario.}
    \label{fig:lidar_point}
\end{figure}

Despite the benefits of using LiDAR data as ground-truth for disparity estimation, challenges arise when capturing outdoor scenes, particularly under adverse weather conditions. Unlike indoor scenes where LiDAR points are densely distributed, outdoor environments often result in sparser point measurements due to various factors. For instance, as shown in Figure~\ref{fig:lidar_point}, outdoor scenes with wet ground caused by rain introduce significant inaccuracies in the LiDAR data. The reflective nature of the wet surface can disrupt the LiDAR signal, leading to incomplete or noisy point measurements. This limitation inhibit the generation of accurate ground-truth disparity maps, especially in regions where the surface is wet or reflective. Figure~\ref{fig:lidar_point} illustrates this issue, where (a) depicts the dual-exposure stereo images of indoor and outdoor scenes, (b) visualizes the disparity map generated by our method, and (c) shows the corresponding LiDAR points. The difference in point density between indoor and outdoor scenes is particularly evident, highlighting the limitations of LiDAR under specific conditions.

\subsection{Initial Exposure Setting}
The initial exposure setting plays a critical role in the performance of dual-exposure disparity estimation. Our exposure control mechanism increases the exposure gap when the scene is determined to have a wide dynamic range, up to a predefined exposure gap. Once this gap is reached, the control mechanism maintains the exposure gap as long as the scene continues to exhibit a wide dynamic range. However, the specific exposure values at which this gap is maintained can vary depending on the initial exposure setting and scene characteristics.

As shown in Figure~\ref{fig:Ablation_Initial_exposure_setting}, when the initial exposures are set to unequal values, the ablation model captures more detail in the first time step due to the larger exposure gap. However, as the exposure gap stabilizes, the model struggles to maintain optimal detail capture, resulting in suboptimal performance compared to the baseline model, which starts with equal exposures. This is particularly evident in scenarios where maintaining the exposure gap is insufficient to fully capture the details of both bright and dark regions.

This observation highlights the importance of carefully selecting the initial exposure setting to balance detail capture across the entire dynamic range of the scene. Future work could focus on adaptive initialization strategies tailored to the scene’s characteristics to improve robustness and consistency.

\bibliographystyle{ACM-Reference-Format}
%\nocite{*}
\bibliography{references}
\end{CJK}